\documentclass{article}

\usepackage{microtype}
\usepackage{graphicx}
\usepackage{subcaption}
\usepackage{booktabs} 

\usepackage{hyperref}

\usepackage[accepted]{icml2026}

\usepackage{amsmath}
\usepackage{amssymb}
\usepackage{mathtools}
\usepackage{amsthm}

\usepackage{enumitem}  

\usepackage{xcolor}
\usepackage{adjustbox}
\usepackage{multirow}
\usepackage{array}
\usepackage{makecell}
\usepackage{bm}

\usepackage{times}
\usepackage{epsfig}
\usepackage{pifont}
\newcommand{\cmark}{\ding{51}}
\newcommand{\xmark}{\ding{55}}

\usepackage{colortbl}
\definecolor{globalcolor}{RGB}{255, 102, 0}
\definecolor{localcolor}{RGB}{0, 153, 76} 
\definecolor{lightgray}{gray}{0.9}

\usepackage{tcolorbox}
\usepackage{listings}
\usepackage{pict2e} 
\usepackage{pgfplots}
\pgfplotsset{compat=1.17}
\usepackage{wrapfig}

\usepackage[normalem]{ulem}   

\newcommand{\revise}[1]{#1}

\usepackage[capitalize,noabbrev]{cleveref}

\theoremstyle{plain}

\theoremstyle{definition}

\theoremstyle{remark}

\icmltitlerunning{Universal Skeleton Understanding via Differentiable Rendering and MLLMs}

\begin{document}

\twocolumn[
  \icmltitle{Universal Skeleton Understanding via Differentiable Rendering and MLLMs}



  \icmlsetsymbol{equal}{*}
  \icmlsetsymbol{corresponding}{$\dagger$}

  \begin{icmlauthorlist}
    \icmlauthor{Ziyi Wang}{pku,equal}
    \icmlauthor{Peiming Li}{pku,equal}
    \icmlauthor{Xinshun Wang}{pku,equal}
    \icmlauthor{Yang Tang}{tencent}
    \icmlauthor{Kai-Kuang Ma}{nuaa}
    \icmlauthor{Mengyuan Liu}{pku,corresponding}
  \end{icmlauthorlist}

  \icmlaffiliation{pku}{State Key Laboratory of General Artificial Intelligence, Peking University, Shenzhen Graduate School, China}
  \icmlaffiliation{tencent}{Tencent}
  \icmlaffiliation{nuaa}{Nanjing University of Aeronautics and Astronautics, Nanjing, Jiangsu, China}

  \icmlcorrespondingauthor{Mengyuan Liu}{liumengyuan@pku.edu.cn}

  \icmlkeywords{Machine Learning, Multimodal Learning, Action Recognition, Skeleton, Differentiable Rendering}

  \vskip 0.3in
]



\printAffiliationsAndNotice{\icmlEqualContribution\textsuperscript{\textdagger}Corresponding author}  

\begin{abstract}
    Multimodal large language models (MLLMs) exhibit strong visual-language reasoning, yet cannot process structured, non-visual data such as human skeletons. Existing methods either compress skeleton dynamics into lossy feature vectors for text alignment, or quantize motion into discrete tokens that generalize poorly across heterogeneous skeleton formats. We present SkeletonLLM, which achieves universal skeleton understanding by translating arbitrary skeleton sequences into the MLLM's native visual modality. At its core is DrAction, a differentiable, format-agnostic renderer that converts skeletal kinematics into compact image sequences. Because the pipeline is end-to-end differentiable, MLLM gradients can directly guide the rendering to produce task-informative visual tokens. To further enhance reasoning capabilities, we introduce a cooperative training strategy: Causal Reasoning Distillation transfers structured, step-by-step reasoning from a teacher model, while Discriminative Finetuning sharpens decision boundaries between confusable actions. SkeletonLLM demonstrates strong generalization \revise{in open-vocabulary action recognition, while its learned reasoning capabilities naturally extend to motion captioning and question answering across heterogeneous skeleton formats}---suggesting a viable path for applying MLLMs to non-native modalities. Code: \url{https://github.com/wangzy01/SkeletonLLM}.
\end{abstract}

\vspace{-6mm}
\section{Introduction}

\begin{figure}[t]
	\centering 
\begin{tabular}{c}		
	\includegraphics[width=\columnwidth]{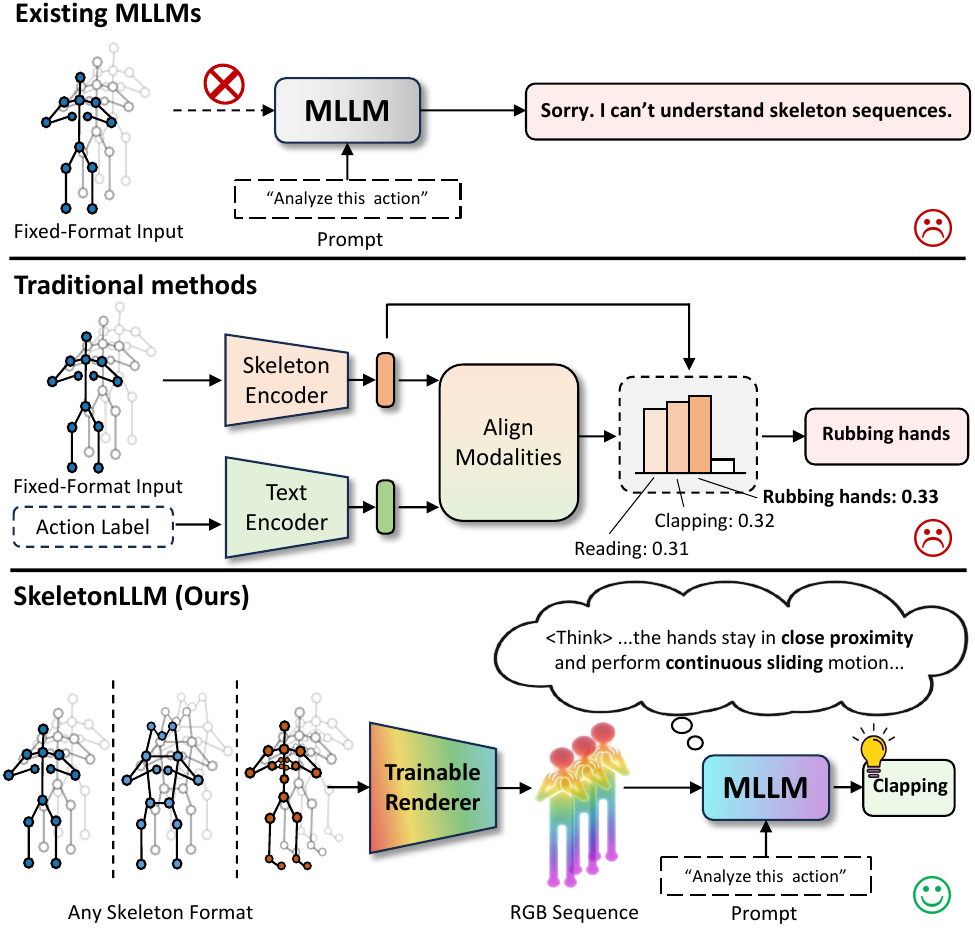}\\
\end{tabular}%
\vspace{-2mm}
	\caption{
        \textbf{Breaking Format Silos and the Modality Gap.}
        \textbf{(Top)} MLLMs possess strong reasoning capabilities but cannot natively process structured skeleton data.
        \textbf{(Middle)} Traditional alignment methods are tied to specific skeleton topologies, compressing motion into a single vector for matching against text embeddings, which creates representation bottlenecks and brittle semantics.
        \textbf{(Bottom)} Our SkeletonLLM uses DrAction, a differentiable renderer that translates a single skeleton sequence of any format into the MLLM's native visual language, enabling end-to-end optimization and unlocking powerful visual reasoning for diverse tasks.
    }\label{fig:fig1}%
    \vskip -0.26in
\end{figure}%
\vspace{-2mm}
Multimodal Large Language Models (MLLMs), such as GPT-5.2 \citep{openai_gpt52_2025} and Qwen-VL \citep{qwen25vl}, have become powerful general-purpose reasoning engines. They show remarkable zero-shot generalization on vision and language tasks. Their success comes from learning rich, transferable representations from massive image-text corpora, enabling them to reason about novel visual concepts using broad world knowledge. A natural question arises: \textit{Can we extend this reasoning capability to structured, non-visual data modalities that MLLMs cannot natively process, thereby achieving universal understanding of such data?}

Human skeleton sequences are a compelling case study. Skeletons offer a compact, privacy-preserving, and appearance-invariant representation of motion \citep{sun2022human}. This makes them attractive for applications where raw video is impractical due to bandwidth, storage, or privacy constraints, such as healthcare and human-robot interaction \citep{wang2025recognizing}. However, skeleton data faces a critical format silo problem. Skeleton topologies vary widely across acquisition systems: Kinect v2 provides 25 joints \citep{zhang2012microsoft}, MoCap systems use 22 SMPL joints \citep{loper2015smpl}, and 2D pose estimators output 17 COCO keypoints \citep{sun2019deep}. This heterogeneity means a model trained on one format cannot be directly applied to another without costly retraining or error-prone joint remapping. Beyond recognition, skeleton-based methods also lack open-ended semantic understanding—they output discrete labels but cannot explain \textit{why} an action occurs or answer natural language questions about fine-grained body dynamics. These limitations—format silos and shallow semantics—have long hindered the goal of universal skeleton understanding: a single model that comprehends any skeleton format across diverse tasks.

Existing approaches to bridge skeletons and foundation models fall into two paradigms, both with fundamental limitations. Feature-text alignment methods \citep{PURLS, STAR, Neuron} train skeleton encoders (e.g., GCNs \citep{yan2018spatial, shi2019two}) to align with text embeddings (e.g., CLIP \citep{radford2021learning}) in a shared latent space. However, these encoders are topology-specific and compress rich spatiotemporal dynamics into a single vector, creating a representation bottleneck that discards details crucial for distinguishing subtle actions. Tokenization-based LLM methods \citep{jiang2023motiongpt, chen2025motionllm} discretize motion into a learned codebook. But this quantization is inherently lossy, and the codebook inherits format dependency—changing the input skeleton structure requires retraining. Critically, both paradigms fail to leverage the MLLM's profound, pre-existing visual understanding. They force the model to learn a ``pose language'' from scratch rather than exploiting its native perceptual capabilities.

We propose \textbf{SkeletonLLM} (\cref{fig:fig1}), a framework that achieves universal skeleton understanding by ``translating'' skeleton sequences into the visual modality that MLLMs natively understand. At its core is \textbf{DrAction} (\textbf{D}ifferentiable \textbf{R}endering of \textbf{Action}s), a learnable renderer that converts skeletal kinematics into image sequences. DrAction is built on 3D Gaussian Splatting \citep{kerbl2023gaussian} and Linear Blend Skinning \citep{loper2015smpl}. It models each pose as deformable Gaussian primitives bound to the kinematic chain. Because the entire pipeline is differentiable, gradients from the MLLM flow back to guide the renderer, enabling it to learn visual representations that are maximally informative for downstream tasks. Crucially, DrAction is format-agnostic: it renders any skeleton—regardless of joint count, connectivity, or coordinate system—into a visually consistent representation, creating a universal visual interface that bypasses the format silo problem entirely.

To strengthen this visual translation, we introduce a cooperative training strategy. \textbf{Causal Reasoning Distillation (CR-Distill)} uses a teacher model to generate step-by-step causal descriptions of body-part dynamics \citep{hinton2015distilling}, guiding the student to learn structured reasoning rather than superficial label mapping. \textbf{Discriminative Finetuning (Disc-FT)} sharpens decision boundaries by training on binary judgments over confusing action pairs (e.g., ``clapping'' vs. ``rubbing hands''). Together, these equip the model with causal understanding and fine-grained discrimination.

We validate SkeletonLLM on diverse tasks: open-vocabulary action recognition on NTU-60/120 \citep{shahroudy2016ntu, liu2019ntu} and PKU-MMD \citep{liu2017pku}, cross-format transfer (Kinect $\leftrightarrow$ MoCap $\leftrightarrow$ 2D poses), motion captioning on HumanML3D \citep{guo2022generating}, and motion question answering. SkeletonLLM substantially outperforms prior methods, with gains amplifying under extreme data scarcity (e.g., +11.96\% on the NTU-60 30/30 split). Notably, models trained on one skeleton format transfer directly to unseen formats without finetuning, demonstrating the format-agnostic nature of our visual translation approach.

\vspace{-1mm}
\textbf{Our contributions are as follows:}
\begin{itemize}
\vspace{-3mm}
\item We propose a new paradigm that translates skeletons into the MLLM's native visual language. For the first time, a single framework unifies heterogeneous skeleton formats (e.g., Kinect, MoCap, 2D poses) and supports multi-task understanding (recognition, captioning, reasoning), moving toward universal skeleton understanding.
\vspace{-2mm}
\item We introduce DrAction, the first differentiable, format-agnostic skeleton renderer for MLLMs, enabling task-optimized visual representations via end-to-end optimization.
\vspace{-2mm}
\item We design a cooperative training strategy (CR-Distill + Disc-FT) that instills causal reasoning and discriminative capability, yielding strong generalization to novel actions and unseen formats, and substantially outperforming prior methods.
\end{itemize}

\section{Related Work}

\subsection{Skeleton-based Action Recognition}

Skeleton sequences provide a compact, privacy-preserving, and appearance-invariant representation of human motion, making them attractive for action recognition. Related structured 3D inputs such as point-cloud videos also highlight the importance of robust spatiotemporal modeling \citep{li2025ust}. Graph Convolutional Networks (GCNs) have become the dominant paradigm, with methods such as ST-GCN \citep{yan2018spatial} and CTR-GCN \citep{ctrgcn} achieving strong supervised performance by modeling the spatial and temporal dependencies of joints. However, these architectures are intrinsically tied to a specific skeleton topology. A model trained on Kinect v2 data (25 joints) cannot be directly applied to MoCap data (22 SMPL joints) or 2D pose estimations (17 COCO keypoints) without retraining or complex joint remapping. This format silo problem severely limits the deployment of skeleton-based models across heterogeneous data sources. Furthermore, these methods can only predict discrete class labels, lacking the capacity for open-ended semantic understanding, such as describing \textit{why} an action is performed or answering questions about fine-grained body-part dynamics.

To generalize to novel action categories, research has extended to Open-Vocabulary Action Recognition. The dominant paradigm here is feature-text alignment \citep{PURLS, STAR, TDSM, SA-DVAE, Neuron}, where a skeleton encoder is aligned with text embeddings (e.g., CLIP \citep{radford2021learning}) in a shared latent space. While effective, this alignment paradigm suffers from a representation bottleneck: compressing rich spatiotemporal dynamics into a single global feature vector often discards the fine-grained motion details needed to distinguish subtle actions. It also suffers from brittle semantic alignment, where simple text labels produce nearly indistinguishable embeddings for similar actions.

\subsection{LLM/MLLM for Motion Understanding}

The rise of Large Language Models (LLMs) and Multimodal LLMs (MLLMs) has opened new avenues for motion understanding. A prominent approach is to tokenize continuous motion into a discrete ``pose vocabulary'' via VQ-VAE, as in MotionGPT \citep{jiang2023motiongpt} and MotionLLM \citep{chen2025motionllm}. Recent unified motion-text-vision frameworks such as UniMotion \citep{wang2026unimotion} further broaden this line toward both understanding and generation. This tokenization allows LLMs to process motion as a sequence of tokens. However, the tokenization paradigm introduces critical limitations: (1) the quantization process is inherently lossy, discarding subtle kinematic details; (2) the learned codebook generalizes poorly to novel skeleton formats; and (3) it imposes an artificial semantic gap, forcing the model to learn an abstract ``pose language'' instead of leveraging its profound, pre-existing visual understanding. SUGAR \citep{sugar2025} instead adopts a more efficient coordinate-to-language pipeline via visual-motion supervision and Temporal Query Projection. An alternative strategy is to encode skeletons into the embedding space of MLLMs, as explored by SKI models \citep{ski2025}. However, this coordinate-based projection struggles to fully activate the MLLM's powerful pretrained perception of visual patterns and lacks a unified mechanism to natively handle diverse skeleton formats. In contrast, our work introduces a differentiable renderer that translates skeletons into the visual modality that MLLMs natively understand. This allows gradients from the MLLM to directly refine the rendering process, learning a visual representation that is maximally informative for the task at hand.

\subsection{Universal Skeleton Representation}

Achieving a universal skeleton representation that transcends format heterogeneity is a long-standing challenge. Prior attempts include joint remapping heuristics\revise{~\citep{duan2022pyskl}}, zero-padding to a maximum joint count\revise{~\citep{wang2024skeleton}}, or learning format-specific adapters\revise{~\citep{guo2023fsar,wang2025heterogeneous}}. These solutions are either lossy, introduce noise, or require retraining for each new format. Our approach sidesteps these issues entirely. By rendering any skeleton---regardless of its joint count, connectivity, or coordinate system---into a visually consistent image sequence, DrAction creates a format-agnostic visual lingua franca. This allows a single model to seamlessly process data from Kinect, MoCap, 2D pose estimators, or any other source, enabling true cross-format generalization without architectural modifications or retraining.

\vspace{-3mm}
\section{Method}

\begin{figure*}[t]
	\centering 
    \begin{tabular}{c}		
		\includegraphics[width=\textwidth]{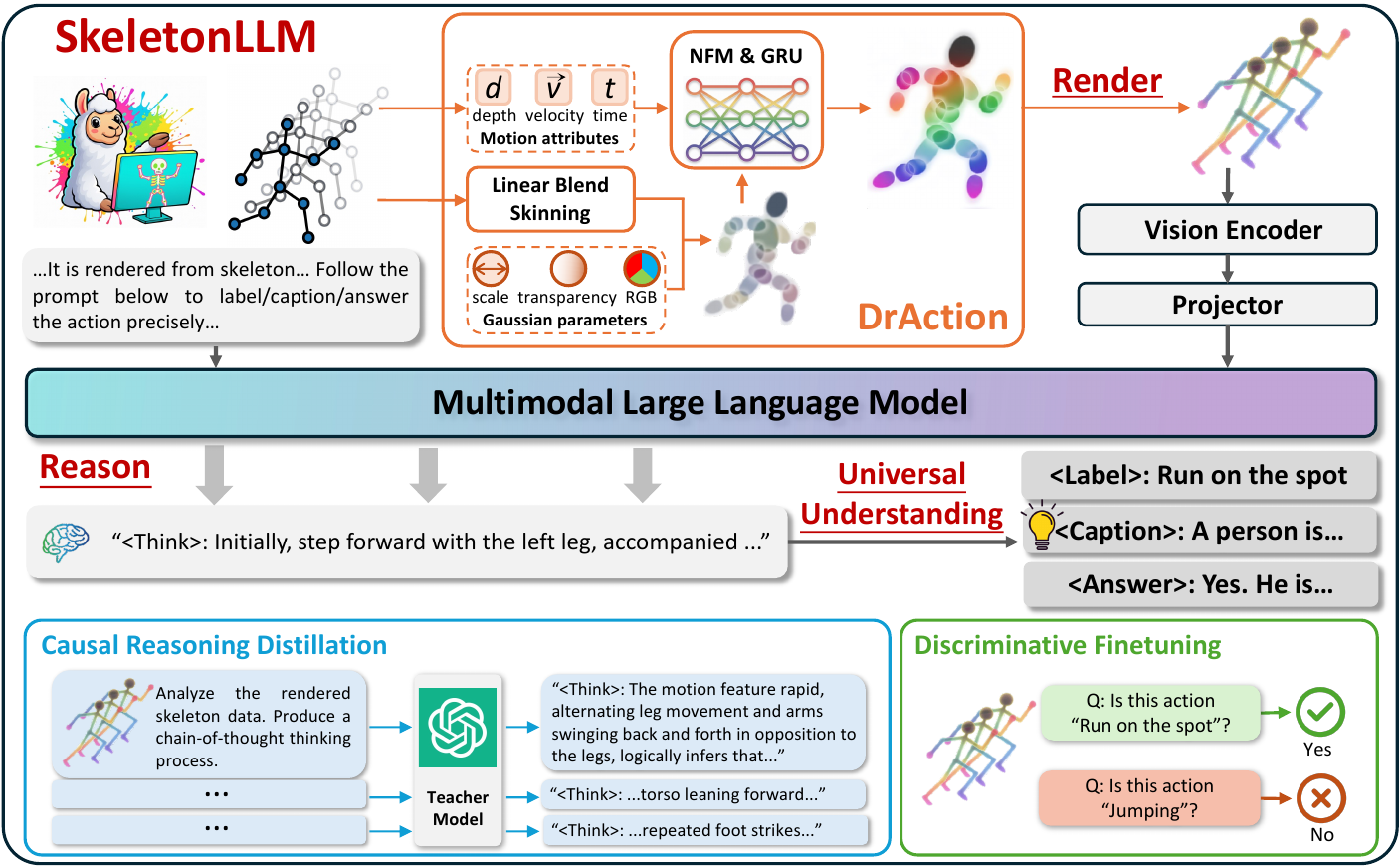}\\
	\end{tabular}%
    \vspace{-3mm}
    \caption{Overview of SkeletonLLM. The pipeline follows a Render-Reason-Respond process for universal understanding. Given a skeleton sequence, DrAction lifts joint trajectories into deformable 3D Gaussian primitives and renders motion-aware images. Joint transforms are computed via Linear Blend Skinning, and kinematic cues (depth, velocity) are fused through a Neural Feature Modulator. All parameters are optimized end-to-end by gradients from the MLLM. The rendered frames are processed by the MLLM's vision encoder and a projector to yield visual tokens. During training, CR-Distill supervises with teacher-generated causal chains describing body-part dynamics, while Disc-FT sharpens decision boundaries via binary queries over confusing action pairs.}\label{fig:framework}%
    \vskip -0.1in
    \vspace{-3mm}
\end{figure*}

\vspace{-2mm}
\subsection{Problem Formulation and Overview}
\vspace{-2mm}
\label{sec:overall_framework}

We address the problem of enabling MLLMs to understand structured, non-visual data outside their native modalities. Specifically, we focus on human skeleton sequences---a modality that is compact, privacy-preserving, and appearance-invariant, yet fundamentally incompatible with the image-text paradigm on which MLLMs are trained.

Let $\mathbf{S} = \{\mathbf{p}_t\}_{t=1}^T$ denote a skeleton sequence, where $\mathbf{p}_t \in \mathbb{R}^{P \times J \times 3}$ represents the 3D coordinates of $J$ joints for $P$ persons at frame $t$. Our goal is to construct a mapping $\mathcal{R}: \mathbf{S} \mapsto \mathbf{V}$ that translates $\mathbf{S}$ into a visual representation $\mathbf{V} = \{\mathbf{I}_t\}_{t=1}^{T'}$ that the MLLM can natively process, while satisfying three desiderata:
\vspace{-1mm}
\begin{enumerate}[leftmargin=*,nosep]
    \item \textbf{Differentiability}: $\mathcal{R}$ must be end-to-end differentiable, allowing gradients from the MLLM's task loss to optimize the translation process.
    \item \textbf{Format Agnosticism}: $\mathcal{R}$ must handle heterogeneous skeleton topologies (varying $J$, different joint definitions, 2D/3D coordinates) without architectural modifications.
    \item \textbf{Information Preservation}: $\mathcal{R}$ must preserve fine-grained spatiotemporal dynamics necessary for distinguishing subtle actions.
\end{enumerate}
\vspace{-1mm}
We realize this mapping through \textbf{DrAction} (\textbf{D}ifferentiable \textbf{R}endering of \textbf{Action}s), a learnable renderer built on 3D Gaussian Splatting and Linear Blend Skinning. As illustrated in \cref{fig:framework}, the complete SkeletonLLM framework operates as a Render-Reason-Respond pipeline: DrAction translates skeleton sequences into image sequences, which are then processed by the MLLM's vision encoder to produce visual tokens for the language model. A cooperative training strategy further enhances reasoning and discriminative capabilities.

\vspace{-3mm}
\subsection{DrAction: Differentiable Skeleton Renderer}
\vspace{-3mm}
\label{sec:3.2}

Existing skeleton visualization methods (e.g., stick figures, heatmaps) are fixed transformations that cannot adapt to downstream tasks. We design DrAction as a learnable visual interface between skeletal kinematics and MLLM perception. The key insight is to model each pose as deformable 3D Gaussian primitives bound to the kinematic chain, enabling both differentiability and expressiveness.

\vspace{-3mm}
\paragraph{Gaussian Primitive Representation}
We model the human body as $K$ deformable 3D Gaussian primitives rather than meshes. To ensure topology invariance, we define these primitives in a canonical, pose-independent space.

For a skeleton with $J$ joints and varying bone connectivity, we instantiate $K = J + K_{\text{bone}}$ Gaussians\revise{, where $K_{\text{bone}} = |\mathcal{E}| \times N_{\text{samples}}$ ($|\mathcal{E}|$: number of bone edges from the skeleton's adjacency list; $N_{\text{samples}}{=}10$: intermediate points per bone)}. The first $J$ Gaussians are anchored at joints, while $K_{\text{bone}}$ Gaussians are uniformly sampled along bone edges via linear interpolation. \revise{Because both $J$ and $|\mathcal{E}|$ are read from the input skeleton, the primitive count adapts automatically to any topology (see Appendix~\ref{sec:format_agnostic_design}).} This density ensures continuous visual connectivity even for sparse skeletons. Each Gaussian $k$ in canonical space is parameterized as:
\vspace{-1mm}
\begin{equation}
    \mathcal{G}_k^c = \{ \boldsymbol{\mu}_k^c, \mathbf{s}_k^c, \mathbf{q}_k^c, \alpha_k^c, \mathbf{f}_k \}
\end{equation}
\vspace{-1mm}
where $\boldsymbol{\mu}_k^c \in \mathbb{R}^3$ is the center, $\mathbf{s}_k^c \in \mathbb{R}^3$ is the scale, $\mathbf{q}_k^c \in \mathbb{R}^4$ is the rotation quaternion, $\alpha_k^c \in [0,1]$ is the base opacity, and $\mathbf{f}_k \in \mathbb{R}^d$ is a learnable appearance feature. Unlike standard 3DGS which optimizes per-scene parameters, here $\boldsymbol{\mu}_k^c$ is initialized from the first frame of each sequence, while $\mathbf{f}_k$\revise{, scales $\mathbf{s}_k^c$, and orientations $\mathbf{q}_k^c$} are learned globally across the dataset.

\vspace{-3mm}
\paragraph{Kinematic Deformation via Linear Blend Skinning}
To animate canonical Gaussians according to input skeletal motion, we employ Linear Blend Skinning (LBS) \citep{loper2015smpl}---a standard technique in computer graphics that provides a principled, differentiable mapping from joint configurations to surface deformations.

\vspace{-1mm}
\textbf{Per-Joint Rigid Transforms.} For each joint $i$ at frame $t$, we compute a rigid transformation $\mathbf{T}_i \in \mathrm{SE}(3)$ that maps from the canonical pose to the current pose. \revise{While \cref{sec:overall_framework} defines a skeleton purely by joint \emph{positions}, some capture systems (e.g., Kinect v2) also record per-joint orientation quaternions that describe the local rotation of the bone segment attached to each joint; these rotations are applied to the Gaussian primitives bound to that joint, not to the joint point itself.} Given the current joint position $\mathbf{j}_i^t$ and orientation quaternion $\mathbf{q}_i^t$ (when available), the transformation is:
\begin{equation}
\mathbf{R}_i = \mathrm{quat2mat}(\mathbf{q}_i^t), \;\; \mathbf{t}_i = \mathbf{j}_i^t - \mathbf{j}_i^c, \;\;
\mathbf{T}_i = \begin{bsmallmatrix} \mathbf{R}_i & \mathbf{t}_i \\ \mathbf{0}^\top & 1 \end{bsmallmatrix}
\end{equation}
where $\mathbf{j}_i^c$ denotes the canonical joint position\revise{---the position of joint $i$ in the first frame of the input sequence, serving as the rest-pose reference for computing per-frame displacements}. When orientation data is unavailable (e.g., for 2D pose estimations \revise{or position-only MoCap}), we \revise{set $\mathbf{R}_i = \mathbf{I}_3$, reducing LBS to} translation-only skinning\revise{---a key mechanism enabling format agnosticism}.

\vspace{-2mm}
\textbf{Blending and SO(3) Projection.} Each Gaussian $k$ is associated with the skeleton via blend weights $\mathbf{w}_k \in \Delta^{J-1}$ (on the probability simplex). \revise{These weights are pre-computed and fixed based on the skeleton's topology: a joint Gaussian at joint $j$ receives a one-hot weight ($w_{k,j}{=}1$); a bone Gaussian interpolated between joints $a$ and $b$ at factor $\alpha$ has $w_{k,a}{=}1{-}\alpha,\; w_{k,b}{=}\alpha$ (zero elsewhere). Thus, the summation in \cref{eq:blend} reduces to at most two non-zero terms for any Gaussian.} The blended translation and rotation are computed as:
\vspace{-1mm}
\begin{equation}
\mathbf{t}_k = \sum_{i=1}^J w_{k,i} \mathbf{t}_i, \qquad \tilde{\mathbf{R}}_k = \sum_{i=1}^J w_{k,i} \mathbf{R}_i
\label{eq:blend}
\end{equation}
\vspace{-3mm}

Since the linear combination $\tilde{\mathbf{R}}_k$ does not generally lie in $\mathrm{SO}(3)$, we project it onto the rotation manifold using polar decomposition via SVD:
\vspace{-1mm}
{\small
\begin{equation}
\tilde{\mathbf{R}}_k = \mathbf{U}\boldsymbol{\Sigma}\mathbf{V}^\top \;\Rightarrow\; \mathbf{R}_k = \mathbf{U} \,\mathrm{diag}(1,1,\det(\mathbf{UV}^\top))\, \mathbf{V}^\top
\end{equation}
}
\vspace{-3mm}

This projection is differentiable and numerically stable, ensuring valid rotations even for extreme poses.

\vspace{-2mm}
\textbf{Gaussian Transformation.} The transformed Gaussian parameters are:
\vspace{-2mm}
\begin{equation}
\boldsymbol{\mu}_k = \mathbf{R}_k \boldsymbol{\mu}_k^c + \mathbf{t}_k, \qquad \mathbf{R}_k^{\text{tot}} = \mathbf{R}_k \mathbf{R}_k^c
\end{equation}
\vspace{-3mm}
\begin{equation}
\boldsymbol{\Sigma}_k = \mathbf{R}_k^{\text{tot}} \cdot \mathrm{diag}\big((\mathbf{s}_k^c)^2\big) \cdot (\mathbf{R}_k^{\text{tot}})^\top
\end{equation}
\vspace{-2mm}
where $\mathbf{R}_k^c = \mathrm{quat2mat}(\mathbf{q}_k^c)$ is the canonical orientation.

\paragraph{Neural Feature Modulator for Pose-Conditioned Appearance}
Static appearance cannot capture the dynamic nature of actions---the same joint configuration may correspond to different motion phases (e.g., arm rising vs. falling). We introduce a \textbf{Neural Feature Modulator (NFM)} that adaptively adjusts each Gaussian's color and opacity based on local kinematics.

\vspace{-1mm}
For each Gaussian $k$, we aggregate position $\mathbf{p}_k^t = \sum_i w_{k,i} \mathbf{j}_i^t$ and velocity $\mathbf{v}_k^t = \sum_i w_{k,i} \dot{\mathbf{j}}_i^t$ \revise{(finite-difference: $\dot{\mathbf{j}}_i^t = \mathbf{j}_i^t - \mathbf{j}_i^{t-1}$)} from associated joints, concatenate with base appearance from \revise{a lightweight} MLP, and process through a \revise{single-layer} GRU for temporal modeling \revise{(it captures motion-phase cues such as acceleration vs.\ deceleration; see Appendix~\ref{sec:ablation_nfm} for ablations on temporal modeling choices)}. The NFM then predicts RGB and opacity residuals $(\Delta\text{RGB}_k, \Delta\alpha_k)$ plus a saliency gate $g_k$. \revise{The gate modulates the final opacity as $\alpha_k{=}\sigma(\alpha_k^{\text{base}}{+}\Delta\alpha_k)\!\cdot\!\sigma(g_k)$, suppressing visually uninformative (e.g., stationary) primitives while amplifying motion-salient ones.} The final color blends the modulated appearance with a depth-based colormap: $\mathbf{C}_k = (1{-}\lambda)\sigma(\text{RGB}_k^{\text{base}} {+} \Delta\text{RGB}_k) + \lambda \mathbf{C}_k^{\text{depth}}$, where \revise{$\sigma(\cdot)$ is the sigmoid function and} $\lambda\revise{{=}\sigma(\theta_{\text{mix}})}$ is a learnable \revise{mixing weight}.

\vspace{-2mm}
\paragraph{Differentiable Rasterization}
We render the transformed Gaussians using the differentiable rasterizer from 3DGS \citep{kerbl2023gaussian}. Each Gaussian is projected onto the image plane via perspective projection:
\vspace{-1mm}
\begin{equation}
\boldsymbol{\mu}_{k,2D} = \pi(\mathbf{K}, \mathbf{W}, \boldsymbol{\mu}_k), \qquad \boldsymbol{\Sigma}_{k,2D} = \mathbf{J} \boldsymbol{\Sigma}_k \mathbf{J}^\top
\end{equation}
\vspace{-1mm}
where $\mathbf{K}$ is the camera intrinsic matrix, $\mathbf{W}$ is the world-to-camera transform, and $\mathbf{J}$ is the projection Jacobian.

The pixel color is computed via front-to-back alpha compositing over depth-sorted Gaussians:
\begin{equation}
\vspace{-1mm}
\mathbf{I}(x, y) = \sum_{k \in \mathcal{N}(x,y)} \mathbf{C}_k \alpha_k' \prod_{j < k} (1 - \alpha_j')
\vspace{-2mm}
\end{equation}
where $\alpha_k' = 1 - \exp(-\alpha_k \cdot w_k(x,y))$ and $w_k(x,y)$ is the Gaussian influence at pixel $(x,y)$.

\vspace{-2mm}
\subsection{Vision-Language Backbone Integration}
\label{sec:3.3}
\vspace{-2mm}

We integrate DrAction as a differentiable front-end to a pre-trained MLLM, forming a unified skeleton-to-language pipeline. The rendered frames $\mathbf{V}$ are processed by the MLLM's vision encoder (ViT) to produce visual tokens, which are projected into the language model's embedding space via a learnable MLP. These tokens replace \texttt{<image>} placeholders in the text prompt, enabling the language model to reason about the depicted motion.

\revise{\textbf{Generative classification via MQA.} Since the MLLM is an auto-regressive generative model, we do \emph{not} add a separate classification head. Instead, action recognition is cast as a \emph{multiple-choice question answering} (MQA) task: the text prompt lists candidate action labels and instructs the model to generate a short answer ending with \texttt{Label: <action>}. The predicted label is extracted via pattern matching at inference. All training losses below are therefore standard auto-regressive cross-entropy on the target token sequence, with prompt tokens masked (see Appendix~\ref{sec:suppl_mqa_prompt} for prompt templates).}

The entire architecture---from skeleton input through rendering, visual encoding, and language generation---is differentiable. This allows the MLLM's task-specific gradients to propagate back through the rendering pipeline, enabling DrAction to learn visual representations that are maximally informative for downstream objectives.

\vspace{-2mm}
\subsection{Cooperative Training Strategy}
\vspace{-2mm}
\label{sec:3.4}

Jointly optimizing a randomly initialized renderer with a pre-trained MLLM presents a ``chicken-and-egg'' dilemma: the MLLM requires recognizable visuals for meaningful supervision, while the renderer needs gradients to learn visualization. We address this via a four-stage cooperative training strategy that progressively instills visual intelligibility, discriminative precision, and causal reasoning.

\vspace{-2mm}
\paragraph{Stage 1: Alignment Warm-up}
We first establish a baseline visual protocol. Keeping the MLLM frozen, we optimize only the DrAction renderer ($\Theta_{\text{render}}$). The objective is a multiple-choice question answering (MQA) task where the model selects the correct action from candidates. This forces the renderer to discover a visual mapping intelligible to the pre-trained vision encoder, aligning skeleton semantics with the MLLM's existing visual priors.

\vspace{-2mm}
\paragraph{Stage 2: Discriminative Finetuning (Disc-FT)}
To address ``brittle semantics'' between similar actions (e.g., ``rubbing hands'' vs. ``clapping''), we introduce a contrastive-like binary classification task. We construct hard negative pairs and ask the model: ``\textit{Does this video show [Action A]?}'' (when it actually shows similar Action B). We update the renderer ($\Theta_{\text{render}}$) and projector ($\Theta_{\text{proj}}$)\revise{; the model generates a single answer token (``Yes''/``No'') and we minimize the auto-regressive negative log-likelihood of this token---equivalently, binary cross-entropy on the positive-token probability---denoted} $\mathcal{L}_{\text{Disc}}$ \revise{(see Appendix~\ref{sec:suppl_discft_prompt})}. This directs the renderer to attend to subtle, discriminative motion details.

\vspace{-2mm}
\paragraph{Stage 3: Causal Reasoning Distillation (CR-Distill)}
Moving beyond recognition to deep understanding, we distill structured reasoning from a stronger teacher model \citep{Hurst2024GPT4oSC}. We prompt the teacher to generate a causal chain of thought: step-by-step analysis of body-part movements, temporal evolution, and potential intent, followed by a conclusion. Our model is trained to generate this complete rationale given the rendered video. We optimize the renderer ($\Theta_{\text{render}}$), projector ($\Theta_{\text{proj}}$), and LLM (via LoRA, $\Theta_{\text{LoRA}}$) using auto-regressive loss $\mathcal{L}_{\text{CR}}$ \revise{over the full teacher token sequence (prompt tokens masked; see Appendix~\ref{sec:suppl_cr_prompt})}. This instills structured, causal action understanding, enabling the model to explain \textit{why} an action is classified as such.

\vspace{-3mm}
\paragraph{Stage 4: Recognition Refinement}
Finally, we freeze the mature renderer and update only the projector ($\Theta_{\text{proj}}$) and LLM ($\Theta_{\text{LoRA}}$) to refine the mapping for open-vocabulary recognition, minimizing \revise{auto-regressive} cross-entropy loss $\mathcal{L}_{\text{MQA}}$ on \revise{the} target label \revise{tokens (same as Stage~1)}.
\vspace{-1mm}

This strategy effectively decouples the complexity of learning to render from learning to reason, ensuring stable convergence and strong generalization.

\begin{table*}[t]
    \scriptsize
    \centering
    \caption{Top-1 accuracy (\%) of various methods on NTU-60 and NTU-120. Each split is denoted as X/Y, where X is the number of seen classes and Y is the number of unseen classes. The best results are in \textbf{\color{red}{red}}, and the second-best are {\color{blue}{\underline{blue}}}. \textsuperscript{\dag}Results for Qwen2.5-VL-7B and InternVL3-8B were obtained by rendering skeletons with \revise{the non-learnable 3D+Velocity renderer (same as in \cref{tab:renderer}; $448{\times}448$)} and finetuning on MQA for 6 epochs.}
    \resizebox{\textwidth}{!}{
    \def\arraystretch{1.2}
    \begin{tabular} {l|c|c|c|c|c|c|c|c|c|c}
        \Xhline{2\arrayrulewidth}
        \multirow{2}{*}{Methods} & \multirow{2}{*}{Type} & \multirow{2}{*}{Venue} &\multicolumn{4}{c|}{NTU-60\;(Acc, \%)} & \multicolumn{4}{c}{NTU-120 \;(Acc, \%)} \\
        \cline{4-11}
        & & & 55/5 & 48/12 & 40/20 & 30/30 & 110/10 & 96/24 & 80/40 & 60/60 \\
        \hline
        ReViSE \citep{ReViSE} & Align & ICCV'17 & 53.91 & 17.49 & 24.26 & 14.81 & 55.04 & 32.38 & 19.47 & 8.27 \\
        JPoSE \citep{JpoSEcvpr} & Align & ICCV'19 & 64.82 & 28.75 & 20.05 & 12.39 & 51.93 & 32.44 & 13.71 & 7.65 \\
        CADA-VAE \citep{CADA-VAE} & Align & CVPRW'19 & 76.84 & 28.96 & 16.21 & 11.51 & 59.53 & 35.77 & 10.55 & 5.67 \\
        SynSE \citep{SynSE} & Align & ICIP'21 & 75.81 & 33.30 & 19.85 & 12.00 & 62.69 & 38.70 & 13.64 & 7.73 \\
        PURLS \citep{PURLS} & Align & CVPR'24 & 79.23 & 40.99 & 31.05 & 23.52 & 71.95 & 52.01 & 28.38 & 19.63 \\
        SCoPLe \citep{SCoPLe} & Align & CVPR'25 & 84.10 & 52.96 & - & - & {\color{blue}{\underline{74.53}}} & 52.17 & - & - \\
        TDSM \citep{TDSM} & Align & ICCV'25 & {\color{blue}{\underline{86.49}}} & 56.03 & {\color{blue}{\underline{36.09}}} & 25.88 & 74.15 & {\color{blue}{\underline{65.06}}} & {\color{blue}{\underline{36.95}}} & {\color{blue}{\underline{27.21}}} \\
        \hline
        MotionGPT \citep{jiang2023motiongpt} & LLM & NeurIPS'23 & 29.88 & 15.91 & 12.14 & 8.57 & 31.10 & 20.39 & 12.96 & 5.15  \\
        MotionLLM \citep{chen2025motionllm} & LLM & TPAMI'25 & 50.24 & 26.98 & 21.58 & 16.46 & 49.80 & 33.62 & 16.91 & 11.45  \\
        \hline
        Qwen2.5-VL-7B\textsuperscript{\dag} \citep{qwen25vl} & MLLM & - & 76.08 & 53.70 & 31.76 & 26.95 & 63.25 & 56.86 & 35.17 & 25.12 \\
        InternVL3-8B\textsuperscript{\dag} \citep{internvl3} & MLLM & -  & 79.66 & {\color{blue}{\underline{56.28}}} & 32.04 & {\color{blue}{\underline{28.15}}}  & 67.03& 58.33& 36.06 & 26.48 \\
        \rowcolor{gray!30} \textbf{SkeletonLLM (Ours)} & MLLM & - & \textbf{\color{red}{87.37}} & \textbf{\color{red}{64.72}} & \textbf{\color{red}{46.15}} & \textbf{\color{red}{37.84}} & \textbf{\color{red}{76.05}} & \textbf{\color{red}{67.20}} & \textbf{\color{red}{44.37}} & \textbf{\color{red}{34.94}} \\
        \Xhline{2\arrayrulewidth}
    \end{tabular}} 
  \label{tab:main}
    \vspace{-2mm}
\end{table*}
\section{Experiments}

We design experiments to answer the following questions:
\begin{enumerate}[leftmargin=*,nosep]
    \item \textbf{Generalization to novel actions}: Can SkeletonLLM recognize actions unseen during training, especially under extreme data scarcity?
    \item \textbf{Cross-format transfer}: Can SkeletonLLM generalize across heterogeneous skeleton formats (e.g., Kinect v2 $\to$ MoCap) without retraining?
    \item \textbf{Semantic understanding}: Beyond classification, can SkeletonLLM perform open-ended motion understanding tasks such as captioning and question answering?
    \item \textbf{Component contributions}: What are the individual contributions of DrAction and each training stage?
\end{enumerate}

\subsection{Experimental Setup}
\vspace{-2mm}

\paragraph{Datasets}
We evaluate SkeletonLLM on diverse benchmarks spanning different skeleton formats, acquisition systems, and task types:
\begin{itemize}[leftmargin=*,nosep]
    \item \textbf{NTU-60 \& NTU-120} \citep{shahroudy2016ntu, liu2019ntu}: Two large-scale benchmarks captured with Kinect v2 (25 joints). NTU-60 contains 60 action classes, and NTU-120 extends it to 120 classes. For NTU-60, we adopt the seen/unseen splits from \citet{SynSE} (55/5, 48/12) and the more challenging splits from \citet{PURLS} (40/20, 30/30). For NTU-120, we use 110/10, 96/24, 80/40, and 60/60 splits.
    \item \textbf{NTU-60 (2D)} \citep{duan2022pyskl}: 2D pose estimations (17 joints) obtained from NTU-60 RGB videos using HRNet \citep{sun2019deep}, used for cross-format evaluation from MoCap to 2D poses.
    \item \textbf{NW-UCLA} \citep{wang2014cross}: A dataset captured with Kinect v1 (20 joints), used for cross-format evaluation.
    \item \textbf{HumanML3D} \citep{guo2022generating}: A motion-language dataset based on AMASS MoCap data (22 SMPL joints), used for motion captioning and cross-format evaluation.
\end{itemize}

\vspace{-2mm}
\paragraph{Implementation Details}
SkeletonLLM is built upon InternVL3-8B \citep{internvl3} and trained on 2 NVIDIA H20 GPUs. For skeleton rendering, each joint and bone segment is modeled as a set of 3D Gaussians, with bone segments constructed by uniformly sampling 10 intermediate points along each connection. We sample 12 frames per sequence using a uniform segment-based strategy and render at $448 \times 448$ resolution. The four progressive training stages are trained for 1, 1, 1, and 3 epochs, respectively. We use the AdamW optimizer with a learning rate of $2 \times 10^{-5}$ and LoRA adapters (rank=32, $\alpha$=64) for efficient finetuning.

\vspace{-2mm}
\paragraph{Baselines}
We compare three categories of methods:
\begin{itemize}[leftmargin=*,nosep]
    \item \textbf{Traditional alignment methods}: Methods such as PURLS \citep{PURLS} and TDSM \citep{TDSM} that use skeleton encoders aligned with text embeddings.
    \item \textbf{LLM-based motion understanding}: Methods that tokenize motion via VQ-VAE for LLM processing \citep{jiang2023motiongpt, chen2025motionllm} or encode skeletons into VLMs \citep{ski2025}.
    \item \textbf{Finetuned MLLMs}: Qwen2.5-VL and InternVL3 \revise{rendered with the non-learnable 3D+Velocity renderer (same as in \cref{tab:renderer}; 12 frames, $448{\times}448$, identical front-view camera)}, finetuned on MQA recognition for 6 epochs.
\end{itemize}

\vspace{-2mm}
\subsection{Open-Vocabulary Action Recognition}
\vspace{-2mm}

We first evaluate SkeletonLLM on standard action recognition benchmarks, focusing on its ability to generalize to action classes unseen during training.

\vspace{-2mm}
\paragraph{Main Results on NTU-60 \& NTU-120}
\cref{tab:main} compares SkeletonLLM against state-of-the-art methods on NTU-60 and NTU-120. To rigorously assess generalization, we evaluate not only conventional splits \citep{SynSE} but also extreme splits \citep{PURLS} that sharply reduce the number of seen classes.

\vspace{-2mm}
\textbf{Performance gap widens under data scarcity.} On the standard 55/5 split, SkeletonLLM outperforms the best traditional method (TDSM) by 0.88\%. However, on the extreme 30/30 split, this gap expands to 11.96\%. This trend is consistent on NTU-120 (7.73\% improvement on the 60/60 split). We attribute this to two factors: (1) DrAction learns visual representations optimized for the MLLM rather than fixed encodings, enabling better generalization; (2) the MLLM's pretrained visual knowledge provides a strong inductive bias for understanding novel motion patterns.

\vspace{-2mm}
\textbf{LLM-based methods underperform finetuned MLLMs.} Despite using large language models, tokenization-based methods (MotionGPT, MotionLLM) consistently lag behind traditional alignment methods and our approach. This suggests that quantizing continuous motion into discrete tokens creates an information bottleneck, discarding the fine-grained kinematic details needed to distinguish similar actions. In contrast, our rendering-based approach preserves the full spatiotemporal richness of motion.

\vspace{-2mm}
\textbf{Differentiable rendering is key.} Comparing SkeletonLLM with InternVL3 using fixed rendering reveals the importance of end-to-end optimization. Despite using the same backbone, SkeletonLLM achieves 8.44\% and 9.69\% higher accuracy on the 48/12 and 30/30 splits of NTU-60, respectively. This confirms that gradients from the MLLM guide DrAction to produce visual representations that are maximally informative for the recognition task. 

\vspace{-2mm}
\subsection{Cross-Format Generalization}
\vspace{-2mm}

A limitation of existing methods is their dependence on specific skeleton topologies. Models trained on Kinect data (25 joints) cannot directly process MoCap data (22 joints) without joint remapping or retraining. We investigate whether SkeletonLLM can overcome this ``format silo'' problem.

\begin{figure}[t]
	\centering 
\begin{tabular}{c}		
		\includegraphics[width=\columnwidth]{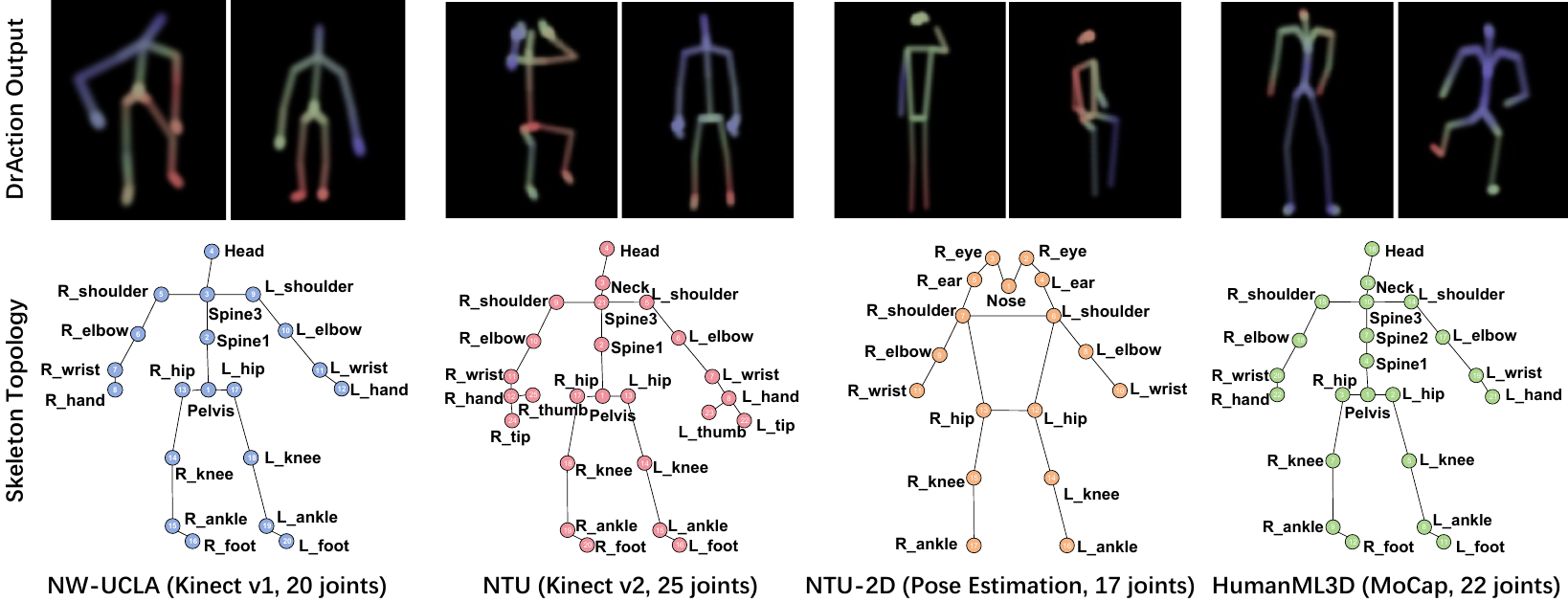}\\
	\end{tabular}%
    \vspace{-2mm}
	\caption{Cross-format rendering by DrAction. Top row: DrAction renders skeletons from four different formats into visually consistent image sequences. Bottom row: the underlying skeleton topologies vary significantly in joint count and connectivity---NW-UCLA (Kinect v1, 20 joints), NTU (Kinect v2, 25 joints), NTU-2D (pose estimation, 17 joints), and HumanML3D (MoCap, 22 joints). Despite these differences, DrAction produces a unified visual language, enabling seamless cross-format transfer.}\label{fig:cross_format}%
    \vskip -0.15in
    \vspace{-1mm}
\end{figure}

\paragraph{Experimental Protocol}
We design cross-format transfer scenarios (NTU-60$\to$NW-UCLA, NTU-60$\to$HumanML3D, HumanML3D$\to$NW-UCLA, and HumanML3D$\to$NTU-60 (2D)) with no finetuning on the target domain. For baselines: MotionGPT uses its expected 263-dim representation after format conversion; SKI-LVLM/TDSM use zero-padding following \citet{wang2024skeleton} and train from scratch.

\begin{table}[t]
\centering
\caption{Cross-format transfer accuracy (\%) for action recognition. Models are trained on the source dataset and evaluated directly on the target without finetuning.}
\label{tab:cross_format_recognition}
\scriptsize
\setlength{\tabcolsep}{3pt}
\renewcommand{\arraystretch}{1.12}
\resizebox{\columnwidth}{!}{%
\begin{tabular}{@{}llllc@{}}
\toprule
Training & Testing & Skeleton Format & Method & Acc \\
\midrule
NTU-60 & NW-UCLA & Kinect v2 $\to$ Kinect v1 & TDSM  & 43.19 \\
NTU-60 & NW-UCLA & Kinect v2 $\to$ Kinect v1 & MotionGPT & 10.35 \\
NTU-60 & NW-UCLA & Kinect v2 $\to$ Kinect v1 & SKI-LVLM  &  31.87 \\
NTU-60 & NW-UCLA & Kinect v2 $\to$ Kinect v1 & \textbf{SkeletonLLM} & \textbf{60.38} \\
\midrule
HumanML3D & NW-UCLA & MoCap $\to$ Kinect v1 & MotionGPT & 9.11\\
HumanML3D & NW-UCLA & MoCap $\to$ Kinect v1 & SKI-LVLM & 28.74 \\
HumanML3D & NW-UCLA & MoCap $\to$ Kinect v1 & \textbf{SkeletonLLM} & \textbf{56.73} \\
\midrule
HumanML3D & NTU-60 (2D) & MoCap $\to$ 2D Pose & MotionGPT &  5.69\\
HumanML3D & NTU-60 (2D) & MoCap $\to$ 2D Pose & SKI-LVLM &  17.13 \\
HumanML3D & NTU-60 (2D) & MoCap $\to$ 2D Pose & \textbf{SkeletonLLM} & \textbf{40.36} \\
\bottomrule
\end{tabular}%
}
\vskip -0.1in
\end{table}

\paragraph{Cross-Format Action Recognition}
\cref{tab:cross_format_recognition} presents results for three cross-format transfer scenarios. When skeleton topology changes, both traditional methods and tokenization-based LLM methods degrade significantly. Zero-padding partially alleviates this but introduces noise that corrupts learned spatial relationships. In contrast, DrAction renders skeletons of different formats into the same visual language---image sequences depicting human motion. As shown in \cref{fig:cross_format}, despite significant differences in skeleton topology (joint count and connectivity), DrAction produces visually consistent renderings. This ``visual lingua franca'' enables motion reasoning regardless of joint configuration. SkeletonLLM achieves 17.19\% and 27.99\% improvements over the best baselines for NTU-60$\to$NW-UCLA and HumanML3D$\to$NW-UCLA transfers, respectively. For HumanML3D$\to$NTU-60 (2D), transferring from 3D MoCap to 2D poses, SkeletonLLM achieves 40.36\% accuracy with a 23.23\% improvement, demonstrating robust generalization even when depth information is entirely absent in the target domain.

\begin{table}[t]
\centering
\caption{Cross-format motion captioning on HumanML3D. Models are trained on NTU-60 (Kinect v2, 25 joints)  with recognition supervision and evaluated on HumanML3D (SMPL, 22 joints) captioning without finetuning.}
\label{tab:cross_ntu_humanml3d}
\scriptsize
\setlength{\tabcolsep}{2.5pt}
\renewcommand{\arraystretch}{1.1}
\resizebox{\columnwidth}{!}{%
\begin{tabular}{@{}lccccccc@{}}
\toprule
Method & R@1$\uparrow$ & R@3$\uparrow$ & MM-Dist$\downarrow$ & BLEU@4$\uparrow$ & ROUGE-L$\uparrow$ & CIDEr$\uparrow$ & BertScore$\uparrow$ \\
\midrule
MotionGPT &2.83 & 9.86 & 12.27 & 3.50 & 13.83 & 4.16 & 14.03  \\
MotionLLM & 4.90 & 14.34 & 9.67 & 5.57 & 19.72 & 5.08 & 23.44 \\
\midrule
InternVL3-8B &6.25 & 20.08 & 6.36 & 7.15 & 27.01 & 9.49 & 29.45   \\
\textbf{Ours} & \textbf{11.60} & \textbf{28.56} & \textbf{5.94} & \textbf{9.63} & \textbf{39.84} & \textbf{18.25} & \textbf{37.28} \\
\bottomrule
\end{tabular}%
}
\vskip -0.1in
\end{table}

\paragraph{NTU-60 $\to$ HumanML3D (Motion Captioning)}
This experiment tests a more challenging scenario: transferring from a recognition-oriented dataset (NTU-60) to a captioning-oriented dataset (HumanML3D), alongside a format shift from Kinect to SMPL. As shown in \cref{tab:cross_ntu_humanml3d}, despite never seeing SMPL-format skeletons or captioning supervision during training, SkeletonLLM demonstrates remarkable transfer capability. This success stems from: (1) DrAction's format-agnostic rendering produces consistent visual representations regardless of joint topology; (2) the CR-Distill stage trains the model to generate descriptive causal chains, which naturally transfers to captioning tasks. SkeletonLLM substantially outperforms both InternVL3-8B (+5.35\% R@1, +8.76 CIDEr over InternVL3-8B) and tokenization-based methods (MotionGPT, MotionLLM), demonstrating that our visual translation approach creates more transferable representations.

\revise{
\paragraph{Single-Model Multi-Task Evaluation}
A key question is whether a \emph{single} model can support all the tasks evaluated above. \cref{tab:single_model} confirms this: the same NTU-60 (55/5) checkpoint---with no task-specific retraining---is used across open-vocabulary recognition (\cref{tab:main}), cross-format recognition (\cref{tab:cross_format_recognition}), cross-format captioning (\cref{tab:cross_ntu_humanml3d}), and motion QA (Appendix Table~15). This demonstrates that SkeletonLLM's visual translation approach yields broadly transferable representations from a single training run.

\begin{table}[t]
\centering
\caption{\revise{Single-model multi-task evaluation. All results use the same NTU-60 (55/5) checkpoint with no task-specific retraining.}}
\label{tab:single_model}
\scriptsize
\setlength{\tabcolsep}{2pt}
\renewcommand{\arraystretch}{1.05}
\resizebox{\columnwidth}{!}{%
\begin{tabular}{@{}llclc@{}}
\toprule
Task & Test Set & Format ($J$) & Superv.? & Metric \\
\midrule
Open-vocab recog. & NTU-60 (55/5) & Kinect v2 (25) & \cmark & 87.37\% Acc \\
Cross-format recog. & NW-UCLA & Kinect v1 (20) & \xmark & 60.38\% Acc \\
Cross-format caption. & HumanML3D & SMPL (22) & \xmark & 37.28 BertScore \\
Motion QA & Skeleton-QA & Kinect v2 (25) & \xmark & 68 / 65\% Acc \\
\bottomrule
\end{tabular}%
}
\vskip -0.1in
\end{table}

\vspace{-2mm}
\paragraph{Cross-Dataset Joint Training}
To further validate the unified framework, we jointly train on NTU-60 (Kinect v2, 25J) and HumanML3D (SMPL, 22J) with shared DrAction and shared MLLM. As shown in \cref{tab:joint_train}, joint training yields consistent positive transfer: NTU-60 recognition improves by 1.15\%, cross-format transfer to NW-UCLA by 4.66\%, and motion QA by 5.2/5.9\%. This confirms that a single model can be jointly trained across heterogeneous datasets, tasks, and skeleton topologies, with DrAction's format-agnostic rendering enabling complementary learning rather than interference.

\begin{table}[t]
\centering
\caption{\revise{Cross-dataset joint training. NTU-60 (25J) + HumanML3D (22J) with shared DrAction and MLLM.}}
\label{tab:joint_train}
\scriptsize
\setlength{\tabcolsep}{4pt}
\renewcommand{\arraystretch}{1.05}
\resizebox{\columnwidth}{!}{%
\begin{tabular}{@{}lccc@{}}
\toprule
Training & NTU-60 55/5 & NTU$\to$NW-UCLA & Skeleton-QA (Temp / Causal) \\
\midrule
NTU-60 only & 87.37 & 60.38 & 68 / 65 \\
Joint training & \textbf{88.52} & \textbf{65.04} & \textbf{73.2} / \textbf{70.9} \\
\bottomrule
\end{tabular}%
}
\vskip -0.1in
\end{table}

\vspace{-2mm}
\paragraph{DrAction with External MLLMs}
Since DrAction outputs standard RGB images, its learned rendering can be directly fed into any closed-source MLLM without modification. \cref{tab:external_mllm} evaluates this: (1) DrAction's rendering improves GPT-4o over fixed 3D+Velocity rendering (+7.78\% on 48/12), confirming that the learned visual representation generalizes beyond the jointly trained backbone; (2) even the powerful GPT-5.4 + DrAction lags behind jointly trained SkeletonLLM by 11.54\% on 48/12, demonstrating that joint end-to-end optimization is irreplaceable and achieves improvements beyond what current frontier LLMs can provide with rendering alone.

\begin{table}[t]
\centering
\caption{\revise{DrAction paired with external closed-source MLLMs on NTU-60 zero-shot recognition. Joint training remains essential.}}
\label{tab:external_mllm}
\scriptsize
\setlength{\tabcolsep}{8pt}
\renewcommand{\arraystretch}{1.05}
\begin{tabular}{@{}lcc@{}}
\toprule
Method & 48/12 & 30/30 \\
\midrule
GPT-4o + 3D+Velocity & 35.74 & 19.27 \\
GPT-4o + DrAction & 43.52 & 25.68 \\
GPT-5.4 + DrAction & 53.18 & 33.11 \\
\textbf{SkeletonLLM (joint-trained)} & \textbf{64.72} & \textbf{37.84} \\
\bottomrule
\end{tabular}
\end{table}
}
\vspace{-2mm}
\subsection{Ablation Study}
\label{sec:ablation}
\vspace{-2mm}

We conduct comprehensive ablations to validate the design choices of SkeletonLLM.

\begin{figure}[t]
	\centering 
	\begin{tabular}{c}		
		\includegraphics[width=\columnwidth]{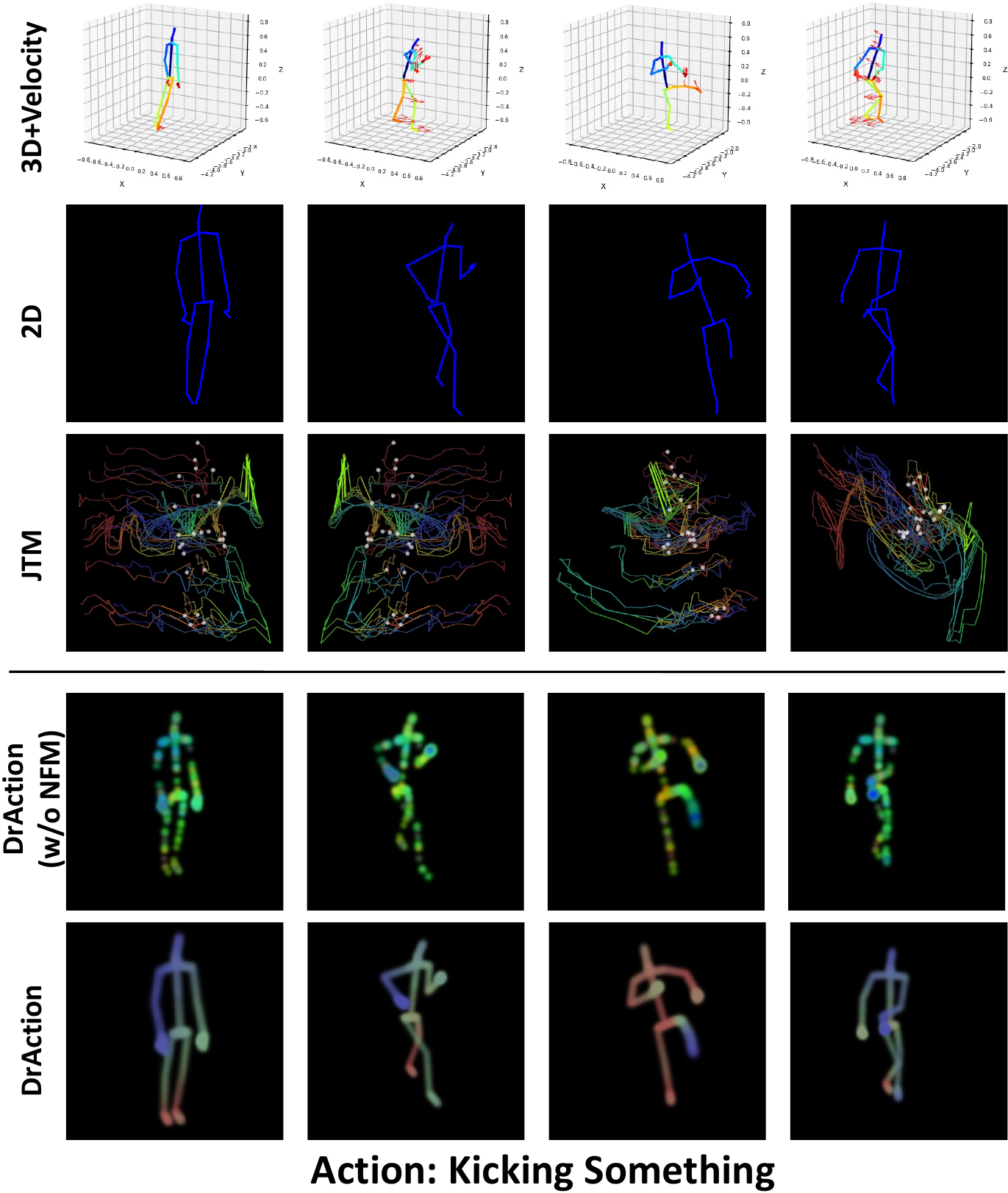}\\
	\end{tabular}%
    \vspace{-2mm}
	\caption{Qualitative comparison of rendering methods. Fixed renderers (3D+Velocity, 2D, JTM \citep{jtmacmmm}) produce visualizations that are either generic, information-poor, or perceptually complex. DrAction learns an abstract representation. With the NFM, it dynamically highlights kinematically salient regions (e.g., the kicking leg), producing a more informative visual language for the MLLM. }\label{fig:fig4}%
    \vskip -0.2in
\end{figure}
\vspace{-3mm}

\vspace{-2mm}
\paragraph{Learnable vs. Non-Learnable Renderers}
\cref{tab:renderer} compares DrAction against fixed, non-learnable renderers. Fixed renderers produce task-agnostic visualizations that may not emphasize kinematically critical regions. JTM's complex multi-view rendering actually hurts performance as the MLLM struggles to interpret its dense visual patterns. Learnable rendering (DrAction w/o NFM) improves over fixed methods by allowing gradients to shape the visual representation. The Neural Feature Modulator (NFM) provides an additional 3.63\% boost on NTU-60 (48/12) by dynamically highlighting motion-salient regions (\cref{fig:fig4}).

\begin{table}[t]
\centering
\caption{Ablation on rendering methods on NTU-60 \& NTU-120. Our differentiable DrAction outperforms non-learnable renderers.}
\label{tab:renderer}
\scriptsize
\setlength{\tabcolsep}{2.5pt}
\renewcommand{\arraystretch}{1.05}
\resizebox{\columnwidth}{!}{%
\begin{tabular}{@{}lcccccccc@{}}
\toprule
\multirow{2}{*}{Renderer} & \multicolumn{4}{c}{NTU-60 (Acc, \%)} & \multicolumn{4}{c}{NTU-120 (Acc, \%)} \\
\cmidrule(r){2-5} \cmidrule(l){6-9}
& 55/5 & 48/12 & 40/20 & 30/30 & 110/10 & 96/24 & 80/40 & 60/60 \\
\midrule
3D+Velocity   & 81.32 & 58.77 & 37.83 & 31.45 & 70.48 & 60.60 & 37.74 & 28.59 \\
2D Projection &81.51 &  58.16 & 38.81 & 31.80 & 70.94 & 61.09 & 38.30 & 27.69 \\
JTM           & 55.86 & 26.37 & 23.45 & 16.83 & 45.51 & 34.03 & 15.09 & 8.75 \\
Ours (w/o NFM)  & 83.30 & 61.09 & 40.93 & 33.87 & 71.29 & 62.62 & 40.69 & 30.67 \\
\textbf{Ours (Full)} & \textbf{87.37} & \textbf{64.72} & \textbf{46.15} & \textbf{37.84} & \textbf{76.05} & \textbf{67.20} & \textbf{44.37} & \textbf{34.94} \\
\bottomrule
\end{tabular}%
}
\vskip -0.1in
\end{table}

\begin{table}[t]
\centering
\caption{Ablation on progressive training strategy on NTU-60 \& NTU-120.}
\vspace{-2mm}
\label{tab:progressive}
\scriptsize
\setlength{\tabcolsep}{2.5pt}
\renewcommand{\arraystretch}{1.05}
\resizebox{\columnwidth}{!}{%
\begin{tabular}{@{}lcccccccc@{}}
\toprule
\multirow{2}{*}{Configuration} & \multicolumn{4}{c}{NTU-60 (Acc, \%)} & \multicolumn{4}{c}{NTU-120 (Acc, \%)} \\
\cmidrule(r){2-5} \cmidrule(l){6-9}
& 55/5 & 48/12 & 40/20 & 30/30 & 110/10 & 96/24 & 80/40 & 60/60 \\
\midrule
w/o CR-Distill & 85.27 & 63.37 & 43.06 & 35.90 & 74.68 & 64.97 & 43.96 & 33.79 \\
w/o Disc-FT & 85.80 & 63.66 & 45.58 & 36.29 & 75.25 & 65.90 & 43.05 & 33.40 \\
w/o Both & 84.43 & 62.09 & 40.78 & 35.05 & 72.95 & 64.05 & 41.13 & 32.77 \\
Joint Training & 82.68 & 61.46 & 40.30 & 34.10 & 72.42 & 62.03 & 40.18 & 30.94 \\
\textbf{Full Pipeline} & \textbf{87.37} & \textbf{64.72} & \textbf{46.15} & \textbf{37.84} & \textbf{76.05} & \textbf{67.20} & \textbf{44.37} & \textbf{34.94} \\
\bottomrule
\end{tabular}%
\vspace{-4mm}
}
\vskip -0.1in
\end{table}
\vspace{-1mm}
\paragraph{Progressive Training Strategy}
\cref{tab:progressive} validates our four-stage progressive training design.

\vspace{-2mm}
\textbf{CR-Distill is critical for reasoning}: Removing it causes the largest drop on extreme splits (1.94\% on 30/30), where causal understanding of motion dynamics matters most. Disc-FT sharpens decision boundaries: removing it particularly affects fine-grained discrimination, as seen from the 1.32\% drop on 80/40. Progressive training prevents optimization interference: jointly training all components from scratch leads to gradient instability and suboptimal convergence, confirming the ``chicken-and-egg'' hypothesis discussed in \cref{sec:3.4}.

\vspace{-2mm}
\section{Conclusion}
\vspace{-2mm}
We introduced SkeletonLLM, a framework that achieves universal skeleton understanding by translating heterogeneous skeleton formats into the MLLM's native visual modality. At its core is DrAction, a format-agnostic, differentiable renderer that converts skeletal kinematics into task-optimized image sequences, enabling end-to-end optimization guided by the MLLM's reasoning gradients. Complemented by a cooperative training strategy that instills causal reasoning and fine-grained discrimination, SkeletonLLM enables a single model to perform recognition, captioning, question answering, and cross-format transfer without architectural modifications. Our results demonstrate that visual translation offers a viable path for extending foundation model reasoning to structured, non-visual data modalities beyond standard images and text.

\section*{Acknowledgements}
This work was supported by National Natural Science Foundation of China (No. 62473007), Guangdong Outstanding Youth Fund (No. 2026B1515020015), Shenzhen Innovation in Science and Technology Foundation for The Excellent Youth Scholars (No. RCYX20231211090248064).

\section*{Impact Statement}
This paper presents work whose goal is to advance the field of Machine Learning. There are many potential societal consequences of our work, none which we feel must be specifically highlighted here.

\bibliography{example_paper}
\bibliographystyle{icml2026}

\newpage
\appendix

\onecolumn
\renewcommand{\thesection}{\Alph{section}}
\renewcommand{\thesubsection}{\thesection.\arabic{subsection}}

\vspace*{1.5cm}
\begin{center}
{\LARGE Appendix Contents}
\vspace{0.8cm}

\begin{tabular}{@{}p{12cm}r@{}}
A. Ablation Studies & \pageref{sec:ablation_studies} \\[6pt]
\quad A.1 Progressive Training Strategy & \pageref{sec:suppl_progressive} \\
\quad A.2 Ablation on Rendering Methods & \pageref{sec:ablation_rendering} \\
\quad A.3 Ablation on NFM Components and Temporal Modeling & \pageref{sec:ablation_nfm} \\[6pt]
B. Additional Experiments & \pageref{sec:additional_exp} \\[6pt]
\quad B.1 Results on PKU-MMD & \pageref{sec:pku_mmd_results} \\
\quad B.2 Ablation on Rendered Frame Count & \pageref{sec:frame_count} \\
\quad B.3 Ablation on Input Resolution & \pageref{sec:resolution} \\
\quad B.4 Error Distribution and Causal Reasoning Analysis & \pageref{sec:error_analysis} \\
\quad B.5 Computational Cost Analysis & \pageref{sec:computational_cost} \\
\quad B.6 Feature Space Analysis & \pageref{sec:feature_analysis} \\[6pt]
C. Cross-Format Generalization & \pageref{sec:cross_format_details} \\[6pt]
\quad C.1 Skeleton Format Differences & \pageref{sec:format_differences} \\
\quad C.2 DrAction's Format-Agnostic Design & \pageref{sec:format_agnostic_design} \\
\quad C.3 Baseline Adaptation for Cross-Format Experiments & \pageref{sec:baseline_adaptation} \\[6pt]
D. Motion Question Answering & \pageref{sec:motion_qa_details} \\[6pt]
\quad D.1 Skeleton-QA Benchmark & \pageref{sec:skeleton_qa} \\
\quad D.2 Benchmark Construction Details & \pageref{sec:benchmark_construction} \\
\quad D.3 Evaluation Protocol & \pageref{sec:eval_protocol} \\
\quad D.4 Quantitative Results & \pageref{sec:qa_results} \\
\quad D.5 Qualitative Analysis & \pageref{sec:qa_qualitative} \\[6pt]
E. Implementation Details & \pageref{sec:implementation_details} \\[6pt]
\quad E.1 Training Setup & \pageref{sec:training_setup} \\
\quad E.2 DrAction Implementation & \pageref{sec:draction_impl} \\
\quad E.3 Details of MQA, Disc-FT and CR-Distill & \pageref{sec:suppl_prompt_design} \\[6pt]
F. Design Rationale: Why 3DGS + LBS & \pageref{sec:design_rationale} \\[6pt]
G. Future Work & \pageref{sec:future_work} \\
\end{tabular}
\end{center}

\clearpage

\twocolumn
\section{Ablation Studies}
\label{sec:ablation_studies}

This section presents comprehensive ablation studies to validate the design choices of SkeletonLLM, including the progressive training strategy, rendering methods, and NFM components.

\subsection{Progressive Training Strategy}
\label{sec:suppl_progressive}

As introduced in the main paper, we propose a progressive four-stage training strategy to address the joint optimization challenge between the differentiable renderer and the MLLM. This section provides a detailed illustration of the training pipeline, including the mathematical formulation of each stage's objective.

\begin{figure*}[!htbp]
	\centering 
	\begin{tabular}{c}		
		\includegraphics[width=\textwidth]{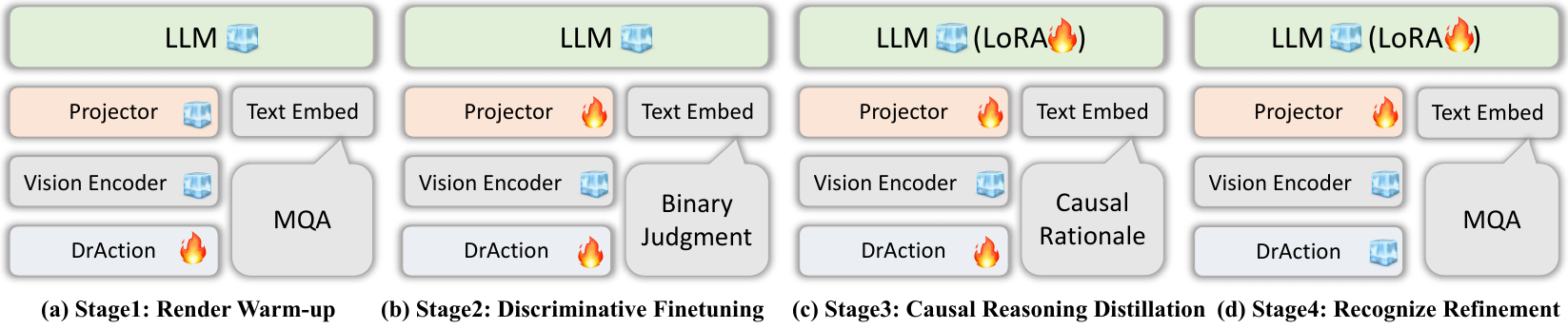}\\
	\end{tabular}%
	\caption{Our Progressive Training Pipeline. To address the joint optimization challenge, we progressively activate and fine-tune model components. The training curriculum begins with (a) warming up the renderer to generate intelligible visuals and concludes with (d) refining recognition, both utilizing a multiple-choice question \& answer (MQA) task. In between, the strategy incorporates (b) learning discriminative features via a binary judgment task and (c) instilling causal reasoning through knowledge distillation from a teacher model.}\label{fig:pipeline_suppl}%
\end{figure*}

\subsubsection{Progressive Training Overview}

\cref{fig:pipeline_suppl} illustrates the complete progressive training pipeline. To address the ``chicken-and-egg'' challenge---where the MLLM requires informative visual inputs to generate meaningful gradients, while the renderer needs such gradients to learn how to produce those visuals---we progressively activate and fine-tune model components in a curriculum manner.

\subsubsection{Stage-wise Training Objectives}

Each stage has a distinct objective, task formulation, and set of trainable parameters. We provide the detailed specification below.

\paragraph{Stage 1: Render Warm-up.}
\begin{itemize}[leftmargin=*,nosep]
    \item Objective: Establish a baseline visual representation that the frozen MLLM can interpret.
    \item Task: Multiple-choice question answering (MQA). Given rendered frames and a list of candidate action labels, the model selects the correct action.
    \item Trainable Parameters: Only DrAction renderer $\Theta_{\text{render}}$; the vision encoder, projector, and LLM remain frozen.
    \item Loss: $\mathcal{L}_{\text{MQA}} = -\log p(y^* | \mathbf{V}, \text{prompt})$, where $y^*$ is the ground-truth label token.
\end{itemize}

\paragraph{Stage 2: Discriminative Finetuning (Disc-FT).}
\begin{itemize}[leftmargin=*,nosep]
    \item Objective: Sharpen decision boundaries between visually similar actions.
    \item Task: Binary judgment on confusing action pairs. For a rendered clip, the model answers whether it depicts a specific (possibly incorrect) action: ``Does this show `rubbing hands'? Yes/No.''
    \item Trainable Parameters: DrAction $\Theta_{\text{render}}$ and the projector MLP $\Theta_{\text{proj}}$.
    \item Loss: $\mathcal{L}_{\text{Disc}} = -\log p(y_{\text{binary}} | \mathbf{V}, \text{query})$, where $y_{\text{binary}} \in \{\text{Yes}, \text{No}\}$.
\end{itemize}

\paragraph{Stage 3: Causal Reasoning Distillation (CR-Distill).}
\begin{itemize}[leftmargin=*,nosep]
    \item Objective: Instill structured, step-by-step reasoning about body-part dynamics.
    \item Task: Knowledge distillation from a teacher model (GPT-4o). The teacher generates detailed causal rationales describing how the action unfolds temporally, grounded in specific body-part movements. The student learns to reproduce these rationales.
    \item Trainable Parameters: DrAction $\Theta_{\text{render}}$, projector $\Theta_{\text{proj}}$, and LLM via LoRA adapters $\Theta_{\text{LoRA}}$.
    \item Loss: $\mathcal{L}_{\text{CR}} = -\sum_{t} \log p(y_t^{\text{teacher}} | y_{<t}^{\text{teacher}}, \mathbf{V}, \text{prompt})$, the standard auto-regressive cross-entropy over the teacher's full token sequence.
\end{itemize}
The teacher rationale follows a structured format: (1) action summary, (2) step-by-step causal chain referencing body parts, (3) final label. This supervision teaches the model not just \textit{what} action occurs, but \textit{why} and \textit{how} it unfolds.

\paragraph{Stage 4: Recognition Refinement.}
\begin{itemize}[leftmargin=*,nosep]
    \item Objective: Consolidate learned capabilities for peak classification accuracy.
    \item Task: Return to MQA, now with a mature renderer and reasoning-aware LLM.
    \item Trainable Parameters: With DrAction frozen, only $\Theta_{\text{proj}}$ and $\Theta_{\text{LoRA}}$ are updated.
    \item Loss: $\mathcal{L}_{\text{MQA}}$ as in Stage 1.
\end{itemize}

\subsubsection{Rationale for Progressive Training}

Our ablations confirm that this curriculum is essential for stable optimization. Joint training from scratch leads to gradient instability and suboptimal convergence due to the warm-start problem. The progressive schedule ensures:
\begin{enumerate}[leftmargin=*,nosep]
    \item The renderer first learns to produce coherent visuals before receiving complex gradients from language generation.
    \item Discriminative training precedes generative reasoning to establish robust feature boundaries.
    \item The final stage consolidates all learned capabilities without disrupting the renderer's learned representations.
\end{enumerate}

\subsubsection{Summary of Component Contributions}

\cref{tab:component_summary} summarizes the role of each component in addressing the challenges of skeleton-MLLM integration.

\begin{table*}[!htbp]
\centering
\caption{Summary of SkeletonLLM components and their contributions.}
\label{tab:component_summary}
\small
\resizebox{\textwidth}{!}{%
\begin{tabular}{lp{5cm}p{6cm}}
\toprule
Component & Addresses & Key Mechanism \\
\midrule
DrAction & Modality gap, format heterogeneity & Differentiable rendering with LBS \\
NFM & Information preservation & Pose-conditioned appearance \\
Cooperative Training & Optimization stability, reasoning depth & Progressive curriculum + distillation \\
\bottomrule
\end{tabular}%
}
\end{table*}

\subsubsection{Quantitative Ablation on Progressive Training}

We conduct a comprehensive ablation study to validate the design of our four-stage progressive training curriculum. Specifically, we evaluate several variants against the full model:
\begin{itemize}
    \item Variant A: Removes Stage 3 (CR-Distill), training without causal reasoning distillation.
    \item Variant B: Removes Stage 2 (Disc-FT), training without discriminative finetuning.
    \item Variant C: Removes both Stage 2 and Stage 3, relying only on render warm-up and recognition refinement.
    \item Variant D: Trains all components jointly from the beginning without a progressive schedule.
\end{itemize}

The quantitative results are presented in \cref{tab:progressive} (main paper). The ablation clearly demonstrates the effectiveness of our progressive training strategy across all evaluation splits on both NTU-60 and NTU-120 datasets. The four variants correspond to: Variant A (w/o CR-Distill), Variant B (w/o Disc-FT), Variant C (w/o Both), and Variant D (Joint Training).

\begin{table*}[!htbp]
\centering
\caption{Ablation on NFM components and temporal modeling strategies on NTU-60 \& NTU-120. We compare different temporal fusion methods (GRU, LSTM, RNN) within the NFM. The full NFM with GRU achieves the best performance across nearly all splits.}
\label{tab:nfm_suppl}
\scriptsize
\renewcommand{\arraystretch}{1.1}
\resizebox{0.8\textwidth}{!}{%
\begin{tabular}{@{}lcccccccc@{}}
\toprule
\multirow{2}{*}{Configuration} & \multicolumn{4}{c}{NTU-60 (Acc, \%)} & \multicolumn{4}{c}{NTU-120 (Acc, \%)} \\
\cmidrule(r){2-5} \cmidrule(l){6-9}
& 55/5 & 48/12 & 40/20 & 30/30 & 110/10 & 96/24 & 80/40 & 60/60 \\
\midrule
w/o NFM & 83.30 & 61.09 & 40.93 & 33.87 & 71.29 & 62.62 & 40.69 & 30.67 \\
w/o Temporal & 86.35 & 63.56 & 45.62 & 36.94 & 74.80 & 65.98 & 43.01 & 33.45\\
w/ RNN & 86.33 & 64.45 & 45.99  & 37.18 & 75.97 & 66.89 & 43.17 & 34.36\\
w/ LSTM  & 86.65 & 64.95 & 45.41 & 37.62 & 75.72 & 67.43 & 43.54 & 34.27 \\
w/ GRU (Full) & 87.37 & 64.72 & 46.15 & 37.84 & 76.05 & 67.20 & 44.37 & 34.94\\
\bottomrule
\end{tabular}%
}
\end{table*}

\paragraph{Analysis of Results.}
The results lead to several important observations:

(1) Both CR-Distill and Disc-FT are critical. Removing either CR-Distill (Variant A) or Disc-FT (Variant B) leads to a significant performance drop across all splits. For instance, on the challenging 30/30 split of NTU-60, removing CR-Distill causes a 1.94\% accuracy drop (from 37.84\% to 35.90\%), while removing Disc-FT results in a 1.55\% drop (to 36.29\%). This confirms that both stages contribute essential and complementary capabilities to the model.

(2) CR-Distill and Disc-FT serve complementary roles. CR-Distill is vital for instilling a deep, causal understanding of complex temporal actions by teaching the model to generate step-by-step reasoning chains that describe body-part dynamics. Disc-FT, on the other hand, is essential for sharpening decision boundaries between visually similar categories by training the model to distinguish confusable action pairs (e.g., ``clapping'' vs. ``rubbing hands'').

(3) The synergistic effect of combining both stages. Variant C, which omits both CR-Distill and Disc-FT, suffers a more substantial degradation than either Variant A or B alone. For example, on the 80/40 split of NTU-120, Variant C achieves only 41.13\%, compared to 43.96\% (Variant A) and 43.05\% (Variant B). This highlights the synergistic and complementary nature of these two training phases.

(4) Progressive training is essential for stable optimization. Variant D, which trains all components jointly from the beginning without a progressive schedule, consistently underperforms compared to the full model across all splits. This confirms our hypothesis that simultaneous end-to-end training is unstable due to the ``chicken-and-egg'' problem. The progressive strategy is crucial for mitigating optimization interference, allowing the renderer to first learn a stable visual representation before the LLM is fine-tuned for complex reasoning tasks. The performance gap is particularly pronounced on more challenging splits with fewer seen classes (e.g., NTU-60 30/30 and NTU-120 60/60), where stable optimization becomes even more critical.

\subsection{Ablation on Rendering Methods}
\label{sec:ablation_rendering}

\cref{tab:renderer} (main paper) compares DrAction against fixed, non-learnable renderers. Fixed renderers produce task-agnostic visualizations that may not emphasize kinematically critical regions. JTM's complex multi-view rendering actually hurts performance as the MLLM struggles to interpret its dense visual patterns. Learnable rendering (DrAction w/o NFM) improves over fixed methods by allowing gradients to shape the visual representation. The Neural Feature Modulator (NFM) provides an additional boost by dynamically highlighting motion-salient regions.

\begin{figure*}[!htbp]
	\centering 
	\begin{tabular}{c}		
		\includegraphics[width=\textwidth]{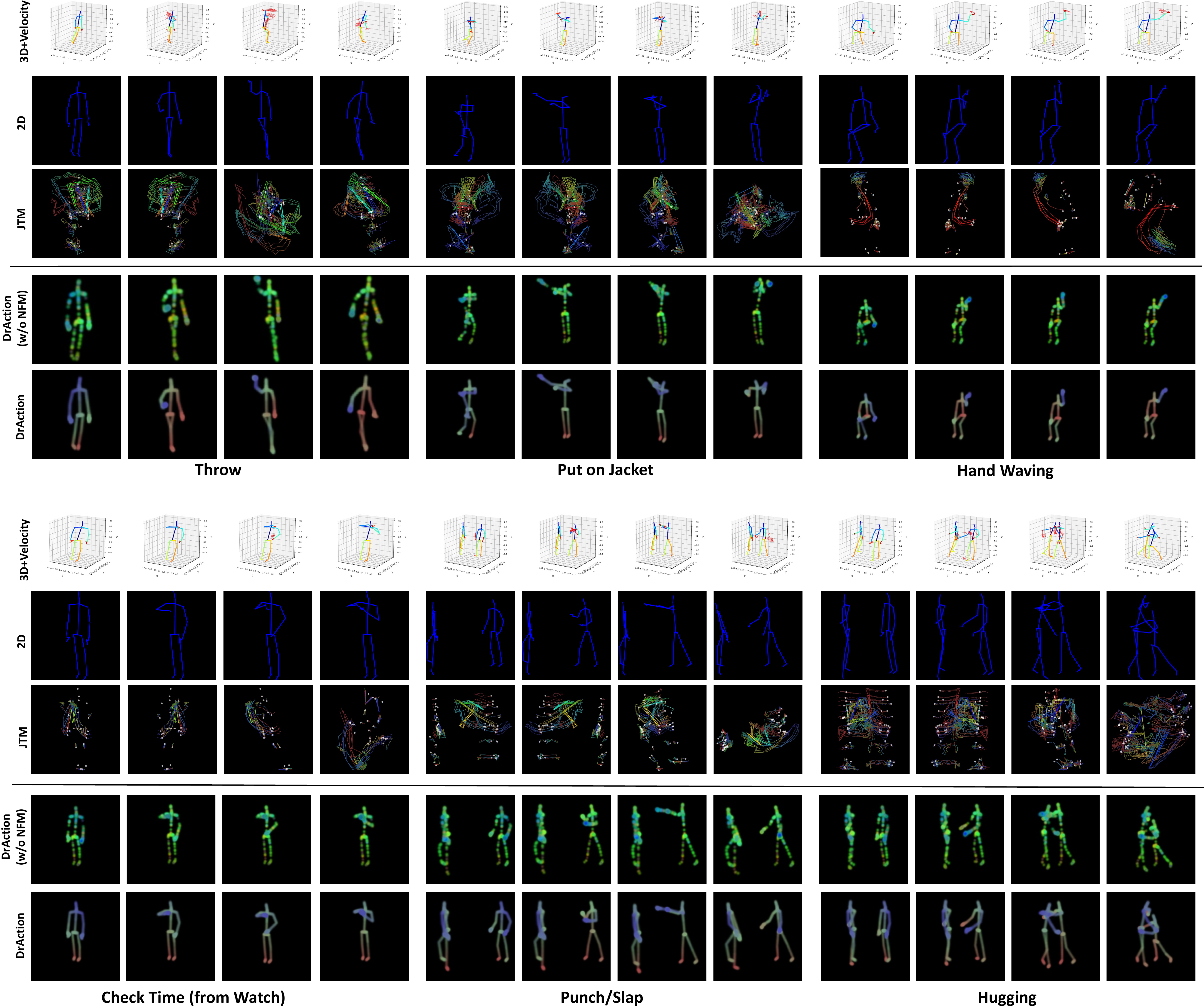}\\
	\end{tabular}%
	\caption{Qualitative comparison of different rendering methods.
 Each column shows a representative action instance.
 From top to bottom, we compare three fixed renderers (3D+Velocity, 2D projection, and JTM~\citep{jtmacmmm}) with our learnable DrAction variants (w/o NFM and full DrAction). Static methods often miss or clutter fine-grained motion cues. In contrast, DrAction, particularly when enhanced by the Neural Feature Modulator (NFM), highlights motion-critical joints using adaptive color and opacity. These visualizations substantiate our claim that a task-optimized, differentiable renderer yields a more effective visual language for MLLMs than static, hand-designed representations. Additional rendered videos are available in the code repository: \url{https://github.com/wangzy01/SkeletonLLM}.}\label{fig:visual}%
\end{figure*}

\cref{fig:visual} provides qualitative comparisons, contrasting our learnable DrAction renderer against fixed rendering pipelines on representative actions. The figure, which organizes actions by column and rendering schemes by row, shows that static methods like 3D+Velocity, 2D projection, and JTM often miss or clutter fine-grained motion cues. In contrast, DrAction, particularly when enhanced by the Neural Feature Modulator (NFM), highlights motion-critical joints using adaptive color and opacity.

\subsection{Ablation on NFM Components and Temporal Modeling}
\label{sec:ablation_nfm}

To validate the contribution of each NFM component, we conduct ablation experiments on NTU-60 and NTU-120. As shown in \cref{tab:nfm_suppl}, removing the NFM entirely (``w/o NFM'') causes a substantial performance drop across all splits, confirming that pose-conditioned appearance modulation is essential for encoding motion-salient cues. 

We also compare different temporal modeling strategies within the NFM, including GRU, LSTM \citep{hochreiter1997long}, and vanilla RNN \citep{elman1990finding}. The GRU component captures temporal dependencies within the feature sequence, enabling the renderer to emphasize phase-specific motion cues (e.g., the acceleration phase of a kick versus the deceleration phase). Among the temporal modeling alternatives, GRU achieves the best performance across most splits, outperforming LSTM by 0.22--0.83\% and vanilla RNN by 0.08--1.20\% on splits where GRU leads. LSTM shows slight advantages on two splits (48/12 on NTU-60 and 96/24 on NTU-120), but introduces additional computational overhead due to its gating mechanism. Vanilla RNN, lacking gating mechanisms, shows competitive results on certain splits but suffers from gradient vanishing issues during training and underperforms on splits requiring longer temporal reasoning. The lightweight GRU strikes the optimal balance between temporal modeling capacity and gradient stability for our per-primitive feature modulation task.

\section{Additional Experiments}
\label{sec:additional_exp}

This section provides additional experimental results, including evaluation on the PKU-MMD dataset, parameter sensitivity analysis on rendered frame count and input resolution, and detailed error analysis with causal reasoning comparisons.

\subsection{Results on PKU-MMD}
\label{sec:pku_mmd_results}

To further assess the generalization of our approach, we evaluate on the PKU-MMD dataset \citep{liu2017pku}, a large-scale multi-modal benchmark with 51 action classes. As shown in \cref{tab:pku_mmd}, for a fair comparison we follow the experimental protocol of \citet{Neuron}, adopting its 46/5 and 39/12 seen/unseen class splits under both the cross-subject (Xsub) and cross-view (Xview) evaluation settings. The results show that SkeletonLLM outperforms all prior methods across both splits and evaluation settings, demonstrating robustness to subject variation and viewpoint changes. Notably, SkeletonLLM also outperforms LLM-based motion understanding methods (MotionGPT, MotionLLM), further validating the effectiveness of our visual translation paradigm over tokenization-based approaches.

\begin{table*}[!htbp]
    \centering
    \caption{Top-1 accuracy (\%) on the PKU-MMD dataset under cross-subject (Xsub) and cross-view (Xview) settings. We follow the splits from \citet{Neuron}. The best results are highlighted in red. \textsuperscript{\dag}Results for Qwen2.5-VL-7B and InternVL3-8B were obtained by rendering skeletons with fixed visualization and finetuning on MQA for 6 epochs.} \label{tab:pku_mmd}
    \scriptsize
    \def\arraystretch{1.2}
    \resizebox{0.8\textwidth}{!}{%
    \begin{tabular}{l|c|c|cc|cc}
        \Xhline{2\arrayrulewidth}
        \multirow{2}{*}{Methods} & \multirow{2}{*}{Type} & \multirow{2}{*}{Venue}
        & \multicolumn{2}{c|}{Xsub (\%)} 
        & \multicolumn{2}{c}{Xview (\%)} \\
        \cline{4-7}
        & & & 46/5 & 39/12 & 46/5 & 39/12 \\
        \hline
        ReViSE \citep{ReViSE} & Align & ICCV'17 & 54.2 & 19.3 & 54.1 & 12.7 \\
        JPoSE \citep{JpoSEcvpr} & Align & ICCV'19 & 57.4 & 27.0 & 53.1 & 22.8 \\
        CADA-VAE \citep{CADA-VAE} & Align & CVPRW'19 & 73.9 & 33.7 & 74.5 & 29.5 \\
        SynSE \citep{SynSE} & Align & ICIP'21 & 69.5 & 36.5 & 71.7 & 25.4 \\
        SMIE \citep{SMIE} & Align & ACM MM'23 & 72.9 & 44.2 & 71.6 & 40.7 \\
        STAR \citep{STAR} & Align & ACM MM'24 & 76.3 & 50.2 & 75.4 & 50.5 \\
        Neuron \citep{Neuron} & Align & CVPR'25 & 89.2 & 61.4 & 88.2 & 62.2 \\
        \hline
        MotionGPT \citep{jiang2023motiongpt} & LLM & NeurIPS'23 & 34.7  & 16.0 & 32.1 & 10.6 \\
        MotionLLM \citep{chen2025motionllm} & LLM & TPAMI'25 & 46.4 & 27.8 & 49.4 & 20.9 \\
        \hline
        Qwen2.5-VL-7B\textsuperscript{\dag} \citep{qwen25vl} & MLLM & - & 80.9 & 57.3 & 83.8 & 60.8 \\
        InternVL3-8B\textsuperscript{\dag} \citep{internvl3} & MLLM & - & 83.4 & 58.2 & 85.0 & 60.6 \\
        \rowcolor{gray!30} SkeletonLLM (Ours) & MLLM & - & {\color{red}{90.1}} & {\color{red}{63.9}} & {\color{red}{89.5}} & {\color{red}{64.2}} \\
        \Xhline{2\arrayrulewidth}
    \end{tabular}%
    }
\end{table*}

\subsection{Ablation on Rendered Frame Count}
\label{sec:frame_count}

We investigate the impact of temporal coverage by varying the rendered frame count $N_{\text{frames}}$ from 4 to 16. As illustrated in \cref{fig:frames_curve}, recognition accuracy consistently improves as more frames are included, since longer sequences capture richer motion dynamics and reduce temporal ambiguity. The performance gain saturates around 12 frames, indicating that most key action phases essential for distinguishing complex actions are sufficiently covered. Consequently, we adopt $N_{\text{frames}}=12$ as the optimal trade-off between recognition performance and computational efficiency.

\begin{figure}[!htbp]
    \centering
    \begin{tikzpicture}[font=\small]
    \begin{axis}[
        xlabel={Number of Frames},
        ylabel={Avg. Acc. on NTU-60 (\%)},
        enlarge x limits=0.1,
        xtick={4, 6, 8, 10, 12, 14, 16},
        xticklabels={4, 6, 8, 10, 12, 14, 16},
        ymin=37, ymax=63,
        ytick={39, 46, 53, 60},
        ymajorgrids=true,
        grid style={dashed, gray!50},
        width=0.95\linewidth,
        height=5cm,
        axis line style={-},
        tick align=outside,
        tickwidth=0.4pt,
        nodes near coords,
        nodes near coords style={font=\scriptsize, color=black, yshift=2pt},
        every node near coord/.append style={
            /pgf/number format/fixed,
            /pgf/number format/precision=2
        }
    ]
    \addplot [
        color=blue,
        mark=*,
        mark options={fill=blue, scale=0.8},
        thick
    ] coordinates {
        (4, 40.10)
        (6, 48.76)
        (8, 54.75)
        (10, 57.94)
        (12, 59.02)
        (14, 59.33)
        (16, 59.45)
    };
    \end{axis}
    \end{tikzpicture}
    \caption{Impact of rendered frame count. Accuracy on NTU-60 increases sharply from 4 to 10 frames and saturates at 12 frames (59.02\%). Increasing to 16 frames yields minimal gains (59.45\%), making 12 frames the optimal trade-off.}
    \label{fig:frames_curve}
\end{figure}

\subsection{Ablation on Input Resolution}
\label{sec:resolution}

We evaluate the impact of rendering resolution ($H \times W$) on recognition performance. As shown in \cref{tab:resolution}, increasing the resolution from $224 \times 224$ to $448 \times 448$ yields a substantial accuracy boost (from 54.24\% to 59.02\%). This significant gain is largely attributed to $448 \times 448$ being the native input resolution of the InternVL3 backbone, which allows the model to extract visual features optimally without resizing artifacts. Although scaling further to $672 \times 672$ provides a marginal improvement (59.98\%) due to finer rasterization details, it incurs disproportionately higher memory and computational costs. We therefore standardize on $448 \times 448$ to achieve the best balance between accuracy and efficiency for all subsequent experiments.

\begin{table}[!htbp]
\centering
\caption{Impact of input resolution. Average Top-1 accuracy on NTU-60 increases with resolution, with $448 \times 448$ offering the best trade-off.}
\label{tab:resolution}
\scriptsize
\resizebox{\columnwidth}{!}{%
\begin{tabular}{lccc}
\toprule
Resolution  & $224 \times 224$ & $448 \times 448$ & $672 \times 672$ \\
\midrule
Top-1 (\%)  & 54.24  & 59.02 & 59.98  \\
\bottomrule
\end{tabular}%
}
\end{table}

\subsection{Error Distribution and Causal Reasoning Analysis}
\label{sec:error_analysis}

\begin{figure*}[!htbp]
	\centering 
	\begin{tabular}{c}		
		\includegraphics[width=\textwidth]{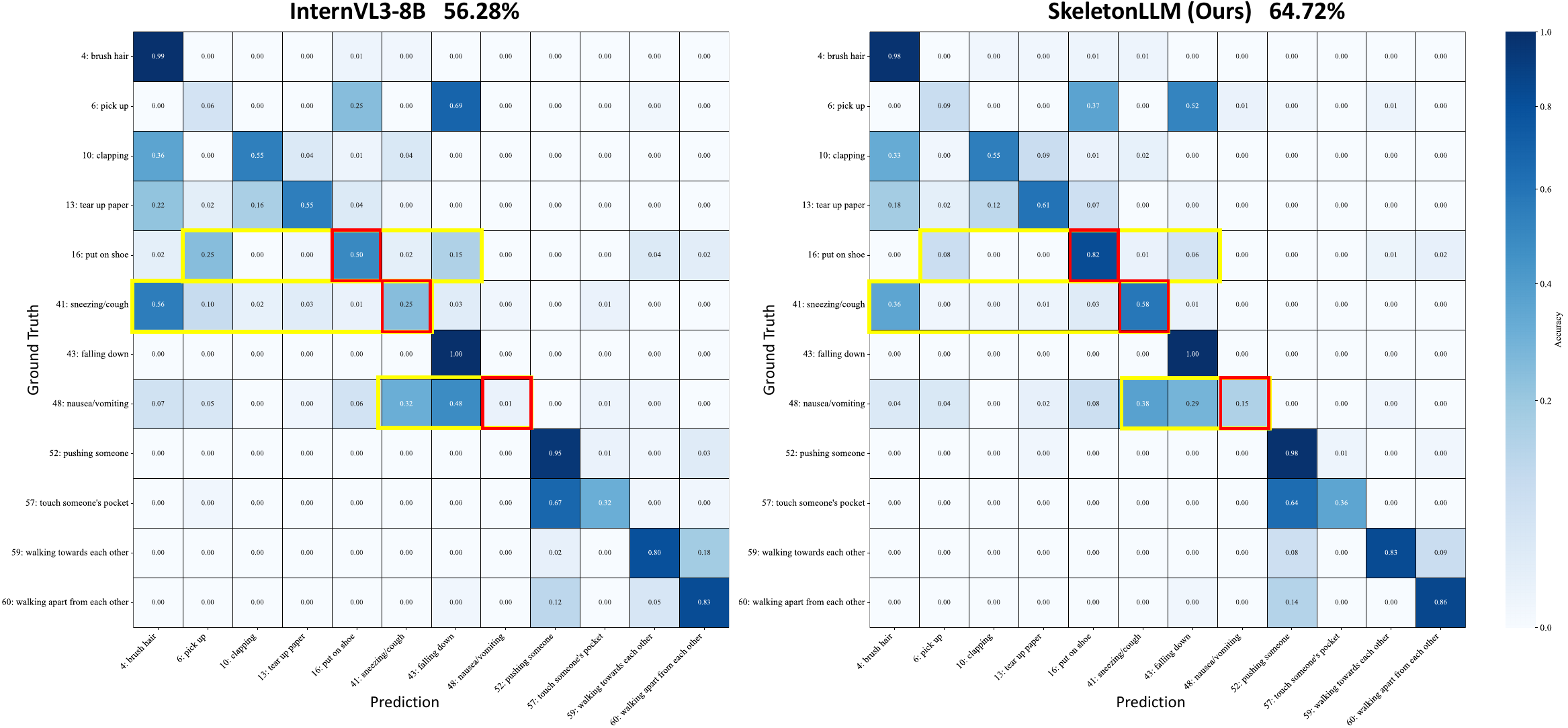}\\
	\end{tabular}%
    \caption{Confusion matrix comparison on the NTU-60 (48/12 split). Left: InternVL3 baseline. Right: SkeletonLLM (Ours). Our method significantly reduces confusion between visually similar actions (highlighted in yellow) and improves accuracy on fine-grained classes (red boxes), demonstrating the effectiveness of Discriminative Finetuning.}\label{fig:heatmap}%
\end{figure*}

\begin{figure*}[!htbp]
	\centering 
	\begin{tabular}{c}		
		\includegraphics[width=\textwidth]{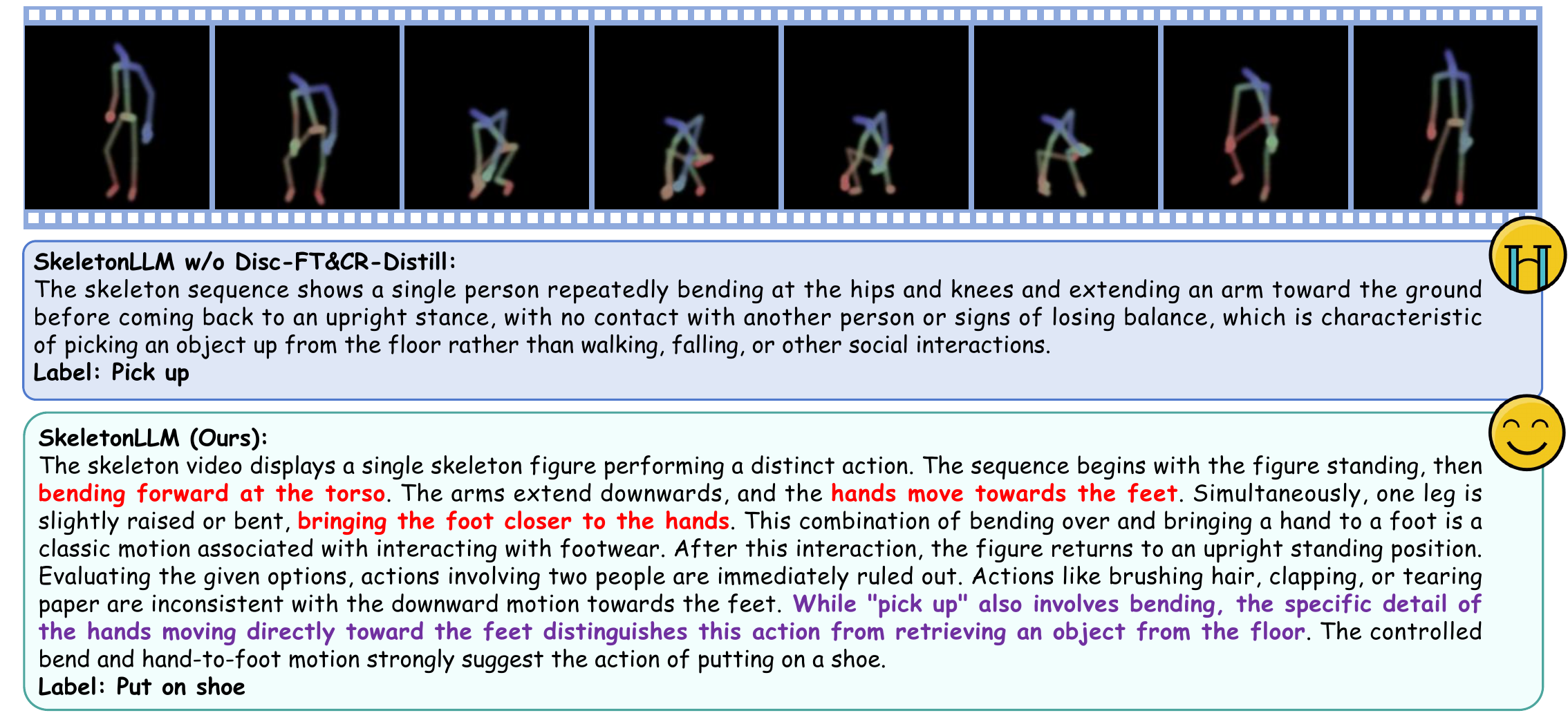}\\
	\end{tabular}%
	\caption{Comparison of reasoning processes on the NTU-60 (48/12 split). We visualize the chain-of-thought generated by SkeletonLLM w/o Disc-FT \& CR-Distill (top) and the full SkeletonLLM (bottom) for a ``Put on shoe'' sequence. While the ablated variant hallucinates a ``Pick up'' action based on a coarse bending posture, the full SkeletonLLM accurately identifies fine-grained hand-foot interactions (red text) and employs causal logic (purple text) to rule out distractors.}\label{fig:cot_cmp}%
\end{figure*}

We provide a qualitative comparison to investigate the source of SkeletonLLM's performance gains on the NTU-60 (48/12 split). \cref{fig:heatmap} visualizes the confusion matrices of the baseline InternVL3 (using fixed rendering and MQA finetuning) versus our SkeletonLLM, while \cref{fig:cot_cmp} contrasts the step-by-step reasoning processes of SkeletonLLM w/o Disc-FT \& CR-Distill and the full SkeletonLLM on a complex action instance.

As shown in \cref{fig:heatmap} (Left), the baseline InternVL3 exhibits significant confusion among visually similar actions. Specifically, within the highlighted yellow regions, distinctive actions like ``put on shoe'' (class 16) are frequently misclassified as ``pick up'' (class 6) or ``falling down'' (class 43), owing to their shared bending postures. Similarly, subtle physiological motions like ``sneezing/cough'' (\cref{fig:heatmap}, class 41) and ``nausea/vomiting'' (class 48) are heavily entangled. In contrast, SkeletonLLM (Right) markedly suppresses these off-diagonal errors. The accuracy for ``put on shoe'' surges from 50\% to 82\%, and the separation between ``sneezing'' and ``nausea'' is substantially clearer (highlighted in red boxes). This improvement validates the effectiveness of our Disc-FT. By explicitly training the model to answer binary ``Yes/No'' questions on these confusable pairs, Disc-FT enforces sharper decision boundaries, preventing the model from relying on superficial pose similarities.

\cref{fig:cot_cmp} further illustrates how the models arrive at their decisions. In the case of ``Put on shoe'', SkeletonLLM w/o Disc-FT \& CR-Distill correctly identifies the ``bending'' posture but fails to discern the fine-grained interaction, hallucinating a ``Pick up'' action. Conversely, the full SkeletonLLM generates a coherent causal chain \citep{wei2022chain}. Although both models employ DrAction, the full model, reinforced by CR-Distill, accurately interprets the visual details, noting that the ``hands move towards the feet'' and the leg is raised to ``bring the foot closer to the hands'' (highlighted in red). Furthermore, driven by this causal reasoning, the model demonstrates robust logic: it explicitly rules out multi-person interactions and distinguishes the hand-to-foot motion from object retrieval (highlighted in purple). This confirms that our cooperative training strategy does not merely fit labels but equips the MLLM with the ability to ground its reasoning in specific, causal kinematic evidence.

\subsection{Computational Cost Analysis}
\label{sec:computational_cost}

To better reflect practical deployment scenarios, we conduct a comprehensive efficiency evaluation of SkeletonLLM on a single consumer-grade NVIDIA RTX 4090 GPU (24GB). All measurements use batch size = 1, with DrAction rendering in FP32 full precision and the MLLM in BF16 mixed precision. We note that the current implementation is an engineering development version without inference-specific optimizations.

\paragraph{DrAction Rendering Efficiency.}
DrAction is built on 3D Gaussian Splatting and benefits from its efficient differentiable rasterization. \cref{tab:render_efficiency} reports rendering latency under different configurations.

\begin{table}[!htbp]
\centering
\caption{DrAction rendering efficiency at different resolutions.}
\label{tab:render_efficiency}
\scriptsize
\resizebox{\columnwidth}{!}{%
\begin{tabular}{lccc}
\toprule
Configuration & Per-frame (ms) & 12 frames (ms) & Memory (GB) \\
\midrule
$448 \times 448$, FP32 & 1.7 & 20.4 & 0.5 \\
$224 \times 224$, FP32 & 1.6 & 19.3 & 0.5 \\
\bottomrule
\end{tabular}%
}
\end{table}

DrAction incurs minimal rendering overhead: rendering 12 frames at $448 \times 448$ takes only $\sim$20ms. Notably, the per-frame latency is nearly identical across resolutions (1.6ms vs. 1.7ms) because 3DGS rendering complexity is dominated by the number of Gaussian primitives rather than output resolution. In our implementation, each skeleton uses a fixed number of primitives ($\sim J + 10 \times |\mathcal{E}|$, where $J$ is joint count and $|\mathcal{E}|$ is the number of bone edges), resulting in stable rendering time regardless of resolution.

\paragraph{End-to-End Inference Performance.}
\cref{tab:e2e_latency} breaks down the complete inference pipeline (skeleton input $\to$ DrAction rendering $\to$ MLLM forward $\to$ text output).

\begin{table}[!htbp]
\centering
\caption{End-to-end inference latency breakdown.}
\vspace{-2mm}
\label{tab:e2e_latency}
\scriptsize
\resizebox{\columnwidth}{!}{%
\begin{tabular}{lcc}
\toprule
Stage & Latency (ms) & Proportion \\
\midrule
DrAction rendering (12 frames) & 20.4 & 10.6\% \\
Vision Encoder & 46.2 & 24.1\% \\
LLM generation (avg. 16 tokens) & 125.0 & 65.3\% \\
\midrule
End-to-end total & 191.6 & 100\% \\
\bottomrule
\end{tabular}%
}
\end{table}
\vspace{-2mm}

The end-to-end latency of $\sim$192ms corresponds to a throughput of $\sim$5.2 samples/sec, meeting real-time interaction requirements ($<$200ms response latency is generally considered the threshold for fluid interaction). The inference bottleneck lies in LLM autoregressive generation ($\sim$65\%), consistent with most MLLM systems where each token requires a full decoder forward pass. The Vision Encoder accounts for $\sim$24\%, processing 12 frames of $448 \times 448$ images ($\sim$3072 visual tokens). DrAction rendering contributes only $\sim$11\% of total compute, indicating that the cost of ``translating'' skeletons to the visual modality is acceptable and does not become a deployment bottleneck.

\paragraph{Memory Footprint Analysis.}
\cref{tab:memory_breakdown} details the memory consumption of each component.

\begin{table}[!htbp]
\centering
\caption{GPU memory breakdown during inference.}
\vspace{-2mm}
\label{tab:memory_breakdown}
\scriptsize
\resizebox{\columnwidth}{!}{%
\begin{tabular}{lcc}
\toprule
Component & Memory (GB) & Note \\
\midrule
InternVL3-8B (BF16) & 13.5 & Model weights \\
LoRA Adapters & 0.4 & Finetuning params \\
DrAction (Gaussians + NFM) & 0.5 & Renderer params \\
Intermediate activations (12 frames) & 1.1 & KV Cache + activations \\
\midrule
Peak inference total & 15.5 & --- \\
\bottomrule
\end{tabular}%
}
\end{table}
\vspace{-2mm}

SkeletonLLM consumes $\sim$15.5GB (65\%) on RTX 4090, leaving ample headroom for larger batch sizes or longer sequences. DrAction occupies only 0.5GB thanks to 3DGS's compact representation. Importantly, DrAction is a lightweight, modular translation module whose functionality is independent of the backend MLLM's scale. This modular design enables flexible combinations: pairing with 1B/2B lightweight MLLMs for low-cost edge deployment, or interfacing with 70B+ models to fully leverage reasoning capacity, providing broad adaptability across different compute platforms.

\vspace{-1mm}
\paragraph{Optimization Potential.}
The reported performance is based on an unoptimized engineering version. In production deployments, mature techniques such as INT8/INT4 weight quantization, KV Cache compression, and speculative decoding can further improve efficiency. With such optimizations, end-to-end latency could potentially drop below 100ms and memory usage to 8--10GB, enabling deployment on a wider range of consumer GPUs.

\vspace{-1mm}
\subsection{Feature Space Analysis}
\label{sec:feature_analysis}
\vspace{-1mm}

\begin{figure}[!htbp]
	\centering 
	\begin{tabular}{c}		
		\includegraphics[width=0.45\textwidth]{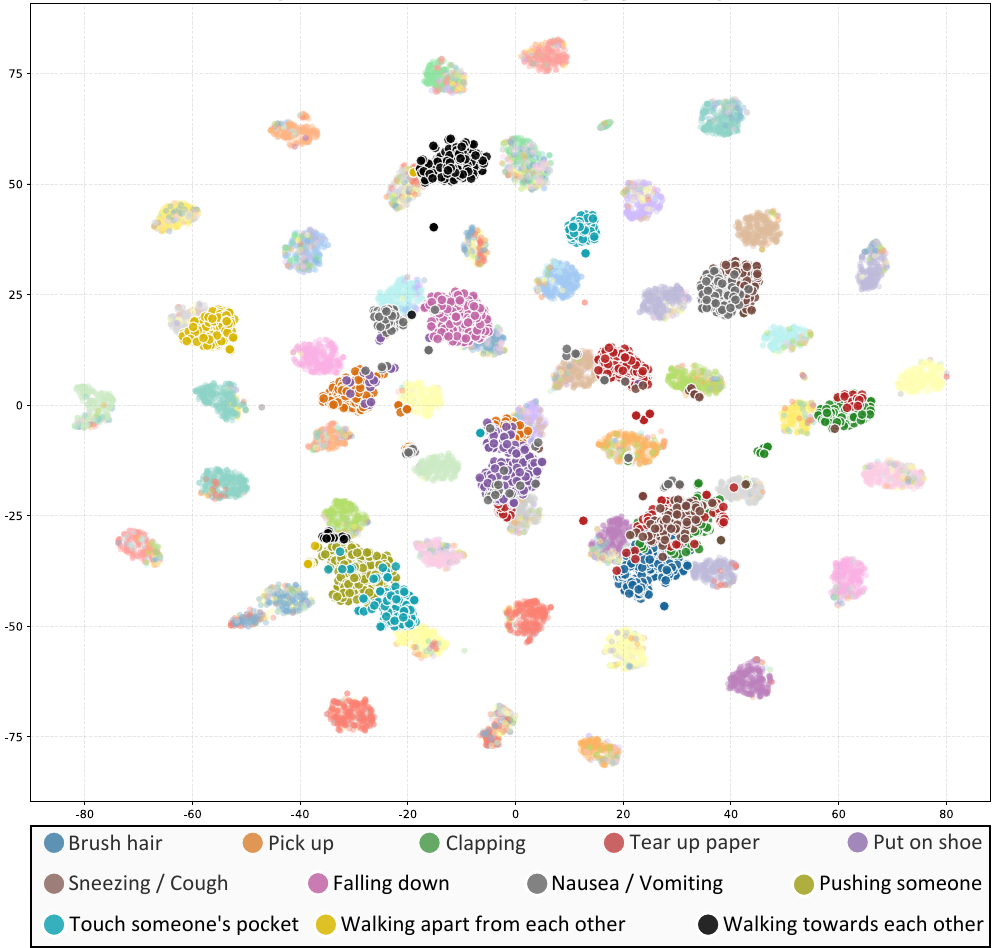}\\
	\end{tabular}%
	\caption{t-SNE visualization of feature representations on NTU-60 (48/12 split). The visualization is derived from the 48/12 open-vocabulary split (Xsub benchmark). Points in faded, lighter colors represent the 48 seen classes, while the darker, highlighted clusters correspond to the 12 unseen classes. Even though the unseen actions were never part of the training set, their samples spontaneously organize into compact, well-separated clusters.}\label{fig:tsne_suppl}%
    \vspace{-5mm}
\end{figure}

We visualize the feature space using t-SNE in \cref{fig:tsne_suppl}. The features are extracted from the hidden states of the MLLM's final Transformer decoder block (before the final normalization and linear head). The visualization corresponds to the Xsub 48/12 open-vocabulary split on NTU-60.

In the plot, the lighter background points represent samples from the 48 seen classes used during training, while the bold, darker points represent the 12 unseen classes encountered only during testing. As shown, even though the unseen actions were never part of the training set, their samples spontaneously organize into compact, well-separated clusters rather than being scattered randomly. This indicates that SkeletonLLM generalizes well, mapping novel motion patterns to distinct semantic regions. Furthermore, visually or semantically similar actions, such as ``Pushing someone'' and ``Touch someone's pocket'', maintain clear separation with distinct boundaries. This suggests that the learned feature space preserves discriminative structure, aiding in the distinction of fine-grained differences among unseen categories, which is essential for robust open-vocabulary recognition.

\vspace{-3mm}
\section{Cross-Format Generalization}
\label{sec:cross_format_details}
\vspace{-3mm}

This section provides additional details on the cross-format transfer experiments presented in the main paper, along with a deeper exposition of DrAction's format-agnostic design principles.

\subsection{Skeleton Format Differences}
\label{sec:format_differences}

\begin{table}[!htbp]
\centering
\caption{Comparison of skeleton formats used in cross-format experiments.}
\label{tab:skeleton_formats}
\scriptsize
\resizebox{\columnwidth}{!}{%
\begin{tabular}{lccc}
\toprule
Dataset & Joints & Acquisition & Topology \\
\midrule
NTU-60/NTU-120 & 25 & Kinect v2 & Kinect skeleton \\
NTU-60 (2D) & 17 & 2D Pose Est. & COCO keypoints \\
NW-UCLA & 20 & Kinect v1 & Kinect skeleton \\
HumanML3D & 22 & MoCap & SMPL model \\
PKU-MMD & 25 & Kinect v2 & Kinect skeleton \\
\bottomrule
\end{tabular}%
}
\end{table}

\cref{tab:skeleton_formats} summarizes the skeleton formats used in our experiments. The key differences include:
\begin{itemize}[leftmargin=*,nosep]
    \item Joint count: Varies from 17 (NTU-60 2D) to 25 (NTU, PKU-MMD), with NW-UCLA at 20 and HumanML3D at 22.
    \item Joint definitions: Kinect and SMPL define joints differently. For example, SMPL includes pelvis as a separate joint while Kinect uses spine base. 2D poses from HRNet follow the COCO keypoint format.
    \item Coordinate systems: Different acquisition systems use different coordinate conventions. Notably, NTU-60 (2D) lacks depth information entirely.
\end{itemize}

\subsection{DrAction's Format-Agnostic Design}
\label{sec:format_agnostic_design}

A critical property of DrAction is its ability to render any skeleton format into a visually consistent representation. This format agnosticism is achieved through three design principles:

\paragraph{Topology-Independent Gaussian Initialization.}
Unlike methods that assume a fixed skeleton structure, DrAction dynamically instantiates Gaussian primitives based on the input skeleton's adjacency structure. The total number of Gaussians adapts automatically:
\begin{equation}
K = J + |\mathcal{E}| \cdot N_{\text{samples}}
\end{equation}
where $J$ is the number of joints, $|\mathcal{E}|$ is the number of bone edges (defined by the skeleton's connectivity), and $N_{\text{samples}}$ is the number of Gaussians sampled per bone (we use $N_{\text{samples}}=10$). Joint Gaussians are centered at joint positions, while bone Gaussians are uniformly interpolated along each edge. This ensures that any skeleton---regardless of its joint count or topology---receives appropriate visual coverage.

\paragraph{Adaptive LBS Weight Computation.}
Linear Blend Skinning weights $\mathbf{w}_k \in \Delta^{J-1}$ are computed based on the input skeleton's connectivity rather than a fixed template:
\begin{itemize}[leftmargin=*,nosep]
    \item Joint Gaussians: Receive concentrated weights via one-hot encoding in logit space, i.e., $\text{logit}_{k,j} = +10$ for the associated joint $j$ and $-10$ otherwise.
    \item Bone Gaussians: For a Gaussian at interpolation factor $\alpha$ between joints $a$ and $b$, the logits are set as $\text{logit}_{k,a} = \log(1-\alpha) + 10$ and $\text{logit}_{k,b} = \log(\alpha) + 10$, with all other entries at $-10$.
\end{itemize}
After softmax normalization, these weights ensure that each Gaussian is primarily influenced by its parent joints, preserving kinematic validity across different skeleton formats.

\paragraph{Invariant Visual Rendering.}
Regardless of the underlying joint count or topology, DrAction produces image sequences depicting human figures in a consistent visual style. This is achieved by:
\begin{enumerate}[leftmargin=*,nosep]
    \item Rendering all skeletons through the same differentiable rasterization pipeline with identical camera parameters.
    \item Using the Neural Feature Modulator (NFM) to produce task-optimized appearances that emphasize motion-salient regions rather than skeleton-specific details.
    \item Blending learned colors with depth-based visualization to maintain spatial coherence across formats.
\end{enumerate}
This creates a \textit{visual lingua franca} that the MLLM can interpret uniformly, enabling seamless cross-format transfer: a model trained on Kinect skeletons (25 joints) can directly process MoCap data (22 joints) or 2D pose estimations (17 joints) without any architectural changes or retraining.

\begin{figure*}[!htbp]
    \centering
    \includegraphics[width=0.88\textwidth]{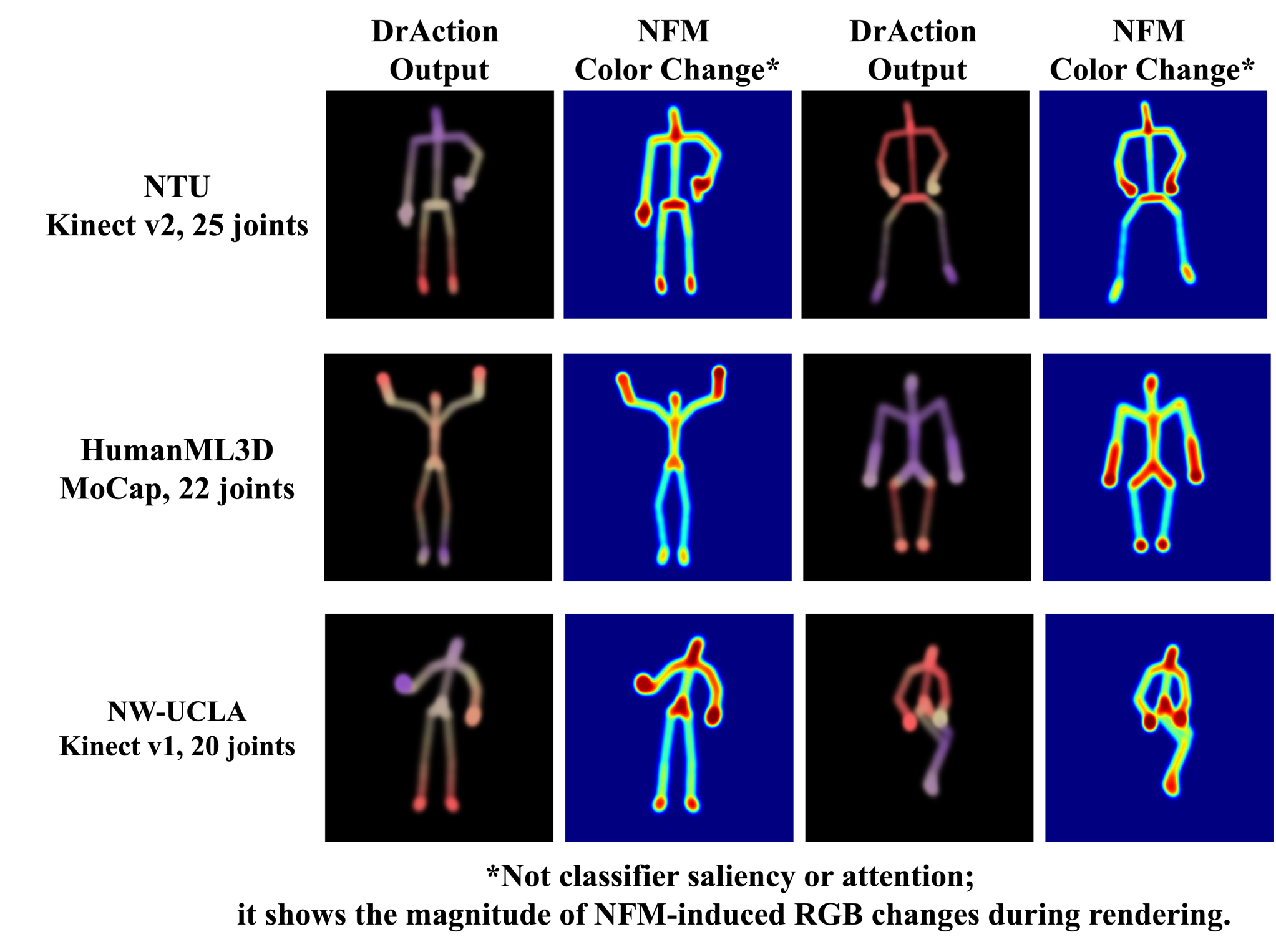}
    \caption{NFM color modulation transfers across skeleton formats.
    We use the same NTU-60-trained DrAction renderer to process examples from three heterogeneous skeleton formats: NTU-60/Kinect v2 (25 joints), HumanML3D/SMPL (22 joints), and NW-UCLA/Kinect v1 (20 joints). Each pair shows the rendered skeleton and the corresponding heatmap of the NFM-induced RGB modification magnitude, where warmer colors indicate stronger appearance updates. Although the joint sets and graph topologies differ, NFM consistently emphasizes semantically corresponding motion-functional regions such as the arms, head-neck area, pelvis, and feet. This suggests that NFM learns local body-part saliency patterns rather than memorizing format-specific joint indices, supporting DrAction's cross-format generalization.}
    \label{fig:nfm_change_cross_format}
\end{figure*}

\paragraph{NFM Behavior Across Formats.}
\cref{fig:nfm_change_cross_format} further visualizes how the Neural Feature Modulator behaves under topology changes. The same learned renderer weights are applied without target-format finetuning, while only the input graph and joint count change. The resulting heatmaps remain concentrated around anatomically and functionally similar regions across formats, indicating that NFM's appearance residuals are driven by local kinematic cues aggregated through sparse topology-based LBS weights. This provides qualitative evidence complementary to the cross-format recognition and captioning results in the main paper.

\subsection{Baseline Adaptation for Cross-Format Experiments}
\label{sec:baseline_adaptation}

For fair comparison, we adapted baseline methods as follows:

MotionGPT: This method expects a fixed 263-dimensional motion representation based on the HumanML3D format. For cross-format experiments:
\begin{itemize}[leftmargin=*,nosep]
    \item We convert source skeletons to the 263-dim representation using joint position, velocity, and rotation features.
    \item Missing joints are interpolated from neighboring joints when possible.
\end{itemize}

SKI-LVLM/TDSM: Following \citet{wang2024skeleton}, we use zero-padding to unify skeleton formats:
\begin{itemize}[leftmargin=*,nosep]
    \item Skeletons are padded to the maximum joint count (25).
    \item Missing joints are filled with zeros.
    \item Models are retrained from scratch on the padded format.
\end{itemize}


\begin{table*}[!htbp]
\centering
\caption{Motion QA performance on the Skeleton-QA benchmark. Temporal, Causal, Fine-grained, and Contrastive columns report MQA accuracy (\%).}
\label{tab:motion_qa}
\scriptsize
\renewcommand{\arraystretch}{1.1}
\resizebox{0.8\textwidth}{!}{%
\begin{tabular}{@{}lcccccc@{}}
\toprule
Method & Temporal & Causal & Fine-grained & Contrastive & ROUGE-L & BertScore \\
\midrule
MotionGPT & 31 & 26 & 33 & 40 & 22 & 34 \\
MotionLLM & 37 & 33 & 39 & 43 & 27 & 39 \\
SKI-LVLM & 29 & 25 & 32 & 38 & 21 & 35 \\
\midrule
InternVL3-8B & 53 & 48 & 55 & 59 & 39 & 49 \\
Ours (w/o CR) & 55 & 52 & 59 & 62 & 43 & 51 \\
Ours (Full) & 68 & 65 & 73 & 76 & 52 & 57 \\
\bottomrule
\end{tabular}%
}
\end{table*}

\section{Motion Question Answering}
\label{sec:motion_qa_details}

Beyond classification and captioning, we investigate whether SkeletonLLM can perform complex reasoning about motion through natural language question answering. This section presents the Motion QA experiments: we first introduce the Skeleton-QA benchmark and describe its construction, then present the evaluation protocol, and finally report quantitative results followed by qualitative analysis.

\subsection{Skeleton-QA Benchmark}
\label{sec:skeleton_qa}

We construct a Motion QA test set based on 200 carefully selected samples from NTU-120, covering four types of reasoning:
\begin{itemize}[leftmargin=*,nosep]
    \item Temporal reasoning: Understanding sequential structure of actions (e.g., ``What is the first step of this action?'')
    \item Causal reasoning: Inferring intentions and consequences (e.g., ``Why might this person be performing this action?'')
    \item Fine-grained understanding: Identifying specific body part movements (e.g., ``Which body parts are primarily involved?'')
    \item Contrastive judgment: Distinguishing similar actions (e.g., ``Is this `clapping' or `rubbing hands'? Explain why.'')
\end{itemize}
Each question has a human-annotated reference answer. We evaluate both MQA (accuracy) and open-ended QA (ROUGE-L, BertScore).

\begin{table}[!htbp]
\centering
\caption{Example question templates for each reasoning type in Skeleton-QA. Templates shown are simplified illustrations; actual prompts include detailed instructions.}
\label{tab:qa_templates}
\scriptsize
\resizebox{\columnwidth}{!}{%
\begin{tabular}{p{1.8cm}p{6cm}}
\toprule
Type & Example Templates \\
\midrule
MQA & ``Analyze the action sequence. Select the best match from options below. Describe your reasoning, then output the final result. A. [action1] B. [action2] ...'' \\
\midrule
Temporal & ``What is the first/last step of this action?'' \\
& ``In what order do the body parts move?'' \\
\midrule
Causal & ``Why might this person be performing this action?'' \\
& ``What could be the goal of this movement?'' \\
\midrule
Fine-grained & ``Which body parts are primarily involved?'' \\
& ``How do the left and right hands move differently?'' \\
\midrule
Contrastive & ``Is this `action A' or `action B'? Explain why.'' \\
& ``What distinguishes this from [similar action]?'' \\
\bottomrule
\end{tabular}%
}
\end{table}

\subsection{Benchmark Construction Details}
\label{sec:benchmark_construction}

The Skeleton-QA benchmark was constructed through the following process:

Sample Selection: We selected 200 samples from and NTU-120, ensuring diversity in:
\begin{itemize}[leftmargin=*,nosep]
    \item Action types (single-person, two-person interactions)
    \item Motion complexity (simple gestures to compound actions)
    \item Temporal length (short clips to long sequences)
\end{itemize}

Question Generation: For each sample, we generated questions across four reasoning types:
\begin{itemize}[leftmargin=*,nosep]
    \item Temporal (1-2 questions): Focus on sequential structure and phase transitions.
    \item Causal (1-2 questions): Focus on intentions, goals, and consequences.
    \item Fine-grained (1-2 questions): Focus on specific body part movements.
    \item Contrastive (1 question): Present similar action pairs for discrimination.
\end{itemize}

\cref{tab:qa_templates} provides example question templates for each reasoning type.

\subsection{Evaluation Protocol}
\label{sec:eval_protocol}

MQA Evaluation: For multiple-choice questions, we provide 4 options (1 correct, 3 distractors). Distractors are selected based on:
\begin{itemize}[leftmargin=*,nosep]
    \item Semantic similarity (e.g., ``clapping'' vs. ``rubbing hands'')
    \item Shared body parts (e.g., both involving arm movements)
    \item Visual similarity in rendered form
\end{itemize}

Open-ended Evaluation: For free-form answers, we compute:
\begin{itemize}[leftmargin=*,nosep]
    \item ROUGE-L: Measures longest common subsequence overlap with reference.
    \item BertScore: Measures semantic similarity using BERT embeddings.
\end{itemize}

\subsection{Quantitative Results}
\label{sec:qa_results}

As shown in \cref{tab:motion_qa}, SkeletonLLM substantially outperforms all baselines across all reasoning types. Based on the rounded scores, the full model improves over the best baseline (InternVL3-8B) by 15 points on temporal reasoning and 17 points on contrastive judgment. The gap between SkeletonLLM (Full) and the ablated version without CR-Distill is 13 points on both temporal and causal reasoning, demonstrating that CR-Distill is critical for instilling structured reasoning capabilities.

\subsection{Qualitative Analysis}
\label{sec:qa_qualitative}

\begin{figure*}[!htbp]
	\centering 
	\begin{tabular}{c}		
		\includegraphics[width=\textwidth]{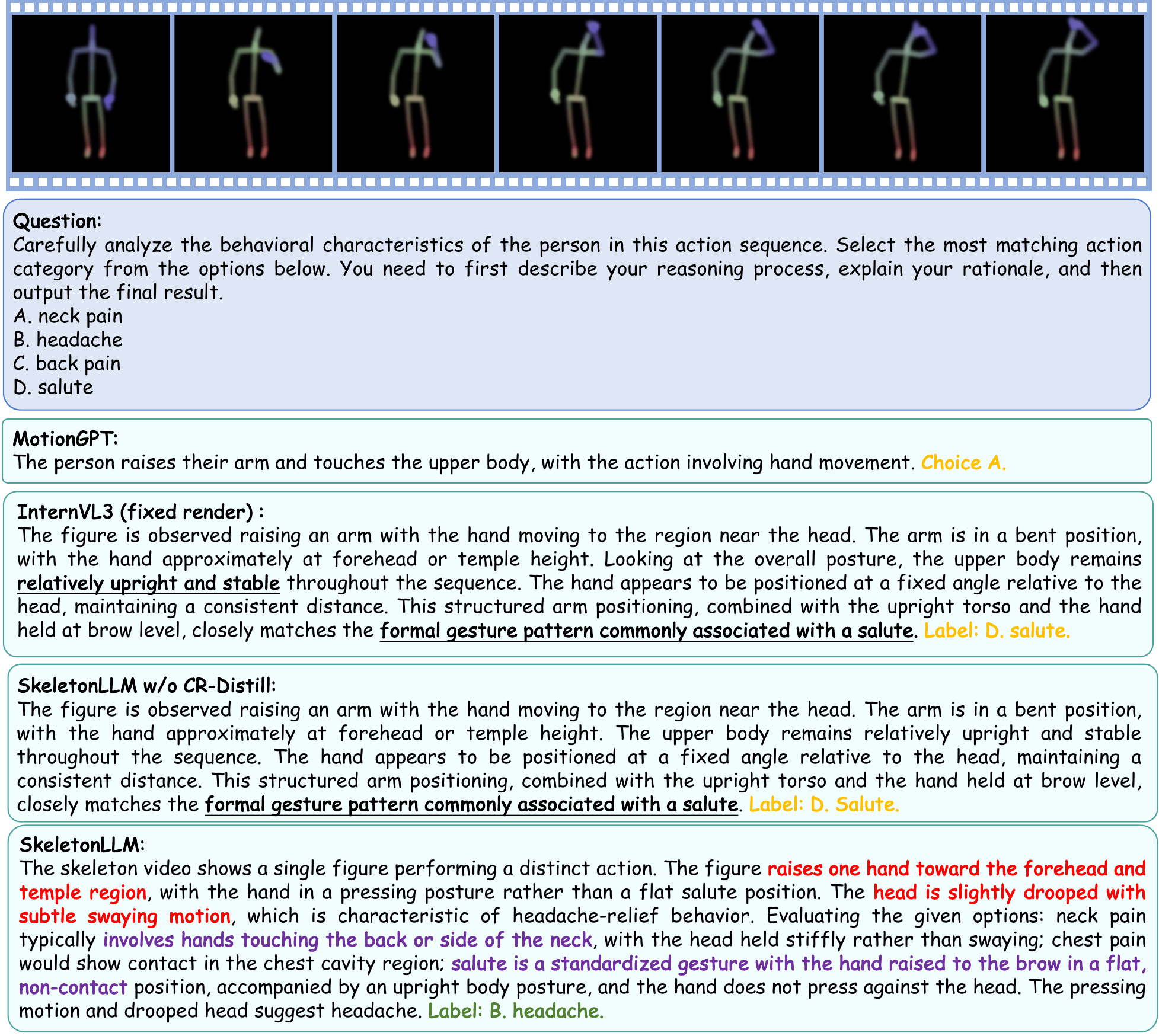}\\
	\end{tabular}%
	\caption{Qualitative comparison of reasoning processes. Given a ``headache'' sequence rendered by DrAction, we compare outputs from different models on a multiple-choice QA task. MotionGPT captures only coarse patterns and selects ``neck pain.'' InternVL3 (fixed render) and SkeletonLLM w/o CR-Distill both misinterpret the gesture as a ``salute,'' focusing on superficial posture cues (``upright and stable,'' ``hand at brow level'') while missing subtle pressing and head-drooping dynamics. Only the full SkeletonLLM correctly identifies ``headache'' by recognizing the pressing posture at the temples and head drooping, while systematically ruling out alternatives with causal reasoning.}\label{fig:motion_qa_example}%
\end{figure*}

\cref{fig:motion_qa_example} illustrates the reasoning processes of different models on a ``headache'' sequence. Given a multiple-choice question asking to identify the action with reasoning, MotionGPT captures only a coarse pattern (``raises their arm and touches the upper body''), failing to distinguish fine-grained motion details and incorrectly selecting ``neck pain.'' InternVL3 with fixed rendering produces a more detailed description, noting the arm movement to the head region with the ``hand approximately at forehead or temple height,'' but critically misinterprets the gesture as a ``salute''---it emphasizes that the upper body remains ``relatively upright and stable'' and describes the ``hand held at brow level'' as matching a ``formal gesture pattern,'' missing the subtle pressing and head-drooping cues. SkeletonLLM without CR-Distill exhibits similar behavior: despite using DrAction's learnable rendering, it still misclassifies the action as ``salute,'' describing a ``formal gesture pattern'' with an ``upright torso.'' This reveals that without causal reasoning supervision, even improved visual representations are insufficient for fine-grained discrimination. In contrast, the full SkeletonLLM correctly identifies ``headache'' by generating a coherent causal chain: it recognizes that the hand is in a ``pressing posture rather than a flat salute position,'' notes the ``head is slightly drooped with subtle swaying motion,'' and systematically rules out alternatives---``neck pain typically involves hands touching the back or side of the neck''; ``salute is a standardized gesture with the hand raised to the brow in a flat, non-contact position.'' This comparison demonstrates that CR-Distill is essential for instilling structured causal reasoning, enabling the model to ground its decisions in specific kinematic evidence rather than superficial pose patterns.

\section{Implementation Details}
\label{sec:implementation_details}

This section provides detailed implementation information for reproducibility.

\subsection{Training Setup}
\label{sec:training_setup}

Our SkeletonLLM is built upon InternVL3-8B \citep{internvl3} and trained on 2 NVIDIA H20 GPUs. For skeleton rendering, each joint and bone segment is modeled as a set of 3D Gaussians \citep{kerbl2023gaussian}, where segments are constructed by uniformly sampling 10 intermediate points along the line connecting each joint pair. To balance temporal coverage and GPU memory usage, we sample 12 frames per sequence using a uniform segment-based strategy \citep{wang2016temporal} and render them at a $448 \times 448$ resolution. The camera intrinsics are derived from a 60° field of view in both horizontal and vertical directions. To ensure stable training, we freeze the parameters of the language model and visual backbone, exclusively training a vision-to-language MLP connector and LoRA adapters \citep{hu2022lora}. These adapters, configured with a rank of 32, a scaling factor $\alpha$ of 64, and a dropout rate of 0.05, are injected into the query/key/value/output projections of the self-attention modules and the up/down/gate projections of the MLP blocks. We employ the AdamW \citep{loshchilov2017decoupled} optimizer with a batch size of 2, an initial learning rate of $2 \times 10^{-5}$, a weight decay of 0.05, and a per-step cosine annealing schedule \citep{loshchilov2016sgdr} with a 0.03 warm-up ratio. The four training stages are trained for 1, 1, 1, and 3 epochs, respectively. For evaluation, we conduct a single test run where the predicted label is compared against the ground-truth after standard text normalization (lowercasing and whitespace trimming).

\subsection{DrAction Implementation}
\label{sec:draction_impl}

This section provides a detailed exposition of our differentiable renderer, DrAction, elaborating on the mathematical formulations, network architectures, and initialization strategies.

\subsubsection{Canonical Representation and Initialization}

The foundation of DrAction is a canonical, action-agnostic representation of the human form as a set of $K$ 3D Gaussian primitives. This representation is initialized once and subsequently deformed by skeletal motion.

\paragraph{Canonical Gaussian Placement and Features.} The total number of Gaussians is $K = J + K_\text{bone}$, where $J$ is the number of joints. The first $J$ Gaussians are centered at the canonical joint positions, $\boldsymbol{\mu}_j^c = \mathbf{j}_j^c$ for $j \in \{1, \dots, J\}$. The canonical joint configuration $\{\mathbf{j}_j^c\}_{j=1}^J$ is initialized from the joint positions of the first available frame (e.g., the first frame of a given sequence) and remains fixed thereafter. The remaining $K_\text{bone}$ Gaussians are uniformly sampled along the skeletal bones, defined by anatomically connected joint pairs. For a bone connecting joints $a$ and $b$, we sample $N$ points, such that the $i$-th point is $\boldsymbol{\mu}_k^c = (1-\alpha_i)\mathbf{j}_a^c + \alpha_i\mathbf{j}_b^c$, where $\alpha_i = i/(N+1)$ for $i \in \{1, \dots, N\}$. The canonical orientation for all Gaussians is initialized to the identity quaternion, $\mathbf{q}_k^c = [1, 0, 0, 0]^\top$. The learnable appearance feature vector $\mathbf{f}_k \in \mathbb{R}^d$ for each Gaussian is initialized from a narrow Gaussian distribution, $\mathbf{f}_k \sim \mathcal{N}(\mathbf{0}, 0.01^2 \mathbf{I})$.

\paragraph{Adaptive Canonical Scaling.}
The canonical scale $\mathbf{s}_k^c$ is initialized adaptively based on local bone structure to ensure appropriate coverage. For a joint Gaussian $j$, its scale is proportional to the median length of bones connected to it. For a bone Gaussian $k$, its scale is proportional to the length of the bone it lies on. This is formulated as:
\begin{equation}
s_j = \text{clip}\left( s_\text{base}^\text{joint} \left( \frac{\text{median}(\{L_b | j \in b\})}{L_\text{max}} \right)^\gamma, s_\text{min}^\text{joint}, s_\text{max}^\text{joint} \right)
\end{equation}
where $L_b$ is the length of bone $b$, $L_\text{max}$ is the maximum bone length in the skeleton, and $s_\text{base/min/max}^\text{joint}$ are hyperparameters. A similar rule applies to bone Gaussians. The final scale vector is $\mathbf{s}_k^c = [s_k, s_k, s_k]^\top$.

\paragraph{Linear Blend Skinning (LBS) Weights.} The blend weights $\mathbf{w}_k \in \mathbb{R}^J$ that associate each Gaussian $k$ with the skeleton's joints are pre-computed and fixed. For a joint Gaussian $j$, its weight vector is effectively a one-hot encoding, with a large positive logit for joint $j$ and large negative logits for all other joints. For a bone Gaussian $k$ sampled between joints $a$ and $b$ with interpolation factor $\alpha$, its weights are non-zero only for $a$ and $b$, with logits proportional to $\log(1-\alpha)$ and $\log(\alpha)$ respectively, ensuring that it is primarily influenced by its parent joints. These fixed weights provide a stable kinematic prior.

\subsubsection{Kinematic Deformation and Rendering}

\paragraph{LBS and Projection to SO(3).}
Our kinematic deformation model utilizes Linear Blend Skinning (LBS) with pose-dependent rigid transforms derived from the skeleton's orientation data. For each joint $i$, its current orientation quaternion $\mathbf{q}_i$ is converted into a rotation matrix $\mathbf{R}_i = \text{quat2mat}(\mathbf{q}_i)$. These per-joint rotations are then blended using the LBS weights $\mathbf{w}_k$:
$
\tilde{\mathbf{R}}_k = \sum_{i=1}^J w_{k,i} \mathbf{R}_i.
$
Since the blended matrix $\tilde{\mathbf{R}}_k$ is not guaranteed to be in the special orthogonal group SO(3), it must be projected to the nearest valid rotation. This is robustly achieved via Singular Value Decomposition \citep{golub2013matrix} (SVD). Let the SVD of the blended matrix be $\tilde{\mathbf{R}}_k = \mathbf{U\Sigma V}^\top$. The closest valid rotation is then given by:
\begin{equation}
\mathbf{R}_k^\text{blend} = \mathbf{U} \text{diag}(1, 1, \det(\mathbf{U V}^\top)) \mathbf{V}^\top.
\end{equation}
This projection corrects for potential reflections and ensures numerical stability. The final posed rotation for a Gaussian is $\mathbf{R}_k = \mathbf{R}_k^\text{blend} \mathbf{R}_k^c$, where $\mathbf{R}_k^c$ is the rotation from its canonical quaternion. The posed covariance matrix $\boldsymbol{\Sigma}_k$ is then computed as in Eq.~(5) of the main paper. It is worth noting that this full kinematic model is employed for datasets providing joint orientation, such as NTU-60 and NTU-120. For datasets like PKU-MMD that lack such data, our model gracefully falls back to a translation-only LBS, where each $\mathbf{R}_i$ is treated as an identity matrix.

\paragraph{Camera Projection and Differentiable Rasterization.}
Given camera intrinsics $\mathbf{K}$ and a world-to-camera transform $\mathbf{W}$, the 3D mean $\boldsymbol{\mu}_k$ is projected to 2D coordinates $\boldsymbol{\mu}_{k, \text{2D}}$. The 3D covariance $\boldsymbol{\Sigma}_k$ is projected to a 2D covariance $\boldsymbol{\Sigma}_{k, \text{2D}}$ via the Jacobian $\mathbf{J}$ of the perspective projection: $\boldsymbol{\Sigma}_{k, \text{2D}} = \mathbf{J} \boldsymbol{\Sigma}_k \mathbf{J}^\top$. For each pixel $\mathbf{x}$, the final color $\mathbf{I}(\mathbf{x})$ is computed via front-to-back alpha compositing:
\begin{equation}
\begin{aligned}
    \mathbf{I}(\mathbf{x}) &= \sum_{k=1}^{K'} \mathbf{C}_k \alpha_k' \prod_{j=1}^{k-1} (1 - \alpha_j'), \\
    \text{where} \quad \alpha_k'(\mathbf{x}) &= 1 - \exp(-\alpha_k \cdot w_k(\mathbf{x})), \\
    \text{and} \quad w_k(\mathbf{x}) &= \exp\left(-\frac{1}{2}(\mathbf{x}-\boldsymbol{\mu}_{k, \text{2D}})^\top\boldsymbol{\Sigma}_{k, \text{2D}}^{-1}(\mathbf{x}-\boldsymbol{\mu}_{k, \text{2D}})\right).
\end{aligned}
\end{equation}
Here, Gaussians are pre-sorted by depth, $w_k(\mathbf{x})$ is the 2D Gaussian influence at the pixel, and $\alpha_k$ is the final modulated opacity for Gaussian $k$.

\paragraph{Camera Configuration and Viewpoint Selection.}
In all experiments, DrAction uses a fixed pinhole camera whose parameters are fully determined by the image resolution and a symmetric field of view, and are not learned during training. Let the rendered frame resolution be $H \times W$ and the horizontal and vertical fields of view be $\phi_x$ and $\phi_y$, respectively. The camera intrinsics are defined as
\begin{equation}
\begin{aligned}
    f_x &= \frac{W}{2\tan(\phi_x / 2)}, \quad
    f_y = \frac{H}{2\tan(\phi_y / 2)}, \\
    c_x &= \frac{W}{2}, \quad c_y = \frac{H}{2}, \\
    \mathbf{K} &= 
    \begin{bmatrix}
        f_x & 0   & c_x \\
        0   & f_y & c_y \\
        0   & 0   & 1
    \end{bmatrix}.
\end{aligned}
\end{equation}
In our implementation, we set $\phi_x = \phi_y = 60^\circ$ and $H = W = 448$, so that $\mathbf{K}$ is uniquely determined by these hyperparameters and remains fixed across sequences.

The world-to-camera transform $\mathbf{W} \in \mathrm{SE}(3)$ is chosen as the identity, i.e.,
\begin{equation}
    \mathbf{W} = 
    \begin{bmatrix}
        \mathbf{I}_3 & \mathbf{0} \\
        \mathbf{0}^\top & 1
    \end{bmatrix},
\end{equation}
so that the 3D joint positions and Gaussian centers are expressed directly in the camera coordinate system. For each Gaussian center $\boldsymbol{\mu}_k = (X_k, Y_k, Z_k)^\top$, the corresponding homogeneous camera coordinate is
\begin{equation}
    \tilde{\boldsymbol{\mu}}_k^c = \mathbf{W}
    \begin{bmatrix}
        \boldsymbol{\mu}_k \\ 1
    \end{bmatrix}
    =
    \begin{bmatrix}
        X_k^c \\ Y_k^c \\ Z_k^c \\ 1
    \end{bmatrix},
\end{equation}
and the pixel location $(u_k, v_k)$ is obtained via
\begin{equation}
\begin{aligned}
    \begin{bmatrix}
        u_k \\ v_k \\ 1
    \end{bmatrix}
    &= \mathbf{K}
    \begin{bmatrix}
        X_k^c / Z_k^c \\ Y_k^c / Z_k^c \\ 1
    \end{bmatrix}, \\
    u_k &= f_x \frac{X_k^c}{Z_k^c} + c_x,\quad
    v_k = - f_y \frac{Y_k^c}{Z_k^c} + c_y,
\end{aligned}
\end{equation}
where the negative sign in the vertical direction follows the standard image-coordinate convention with $v$ increasing downward.

Before projection, we apply a simple depth normalization to keep the skeleton comfortably in front of the camera and to avoid numerical issues near $Z_k^c \approx 0$. Let $\{\mathbf{j}_i^t\}_{t,i}$ denote the joints of the entire sequence after LBS. We compute the sequence-level median depth
\begin{equation}
    \tilde{z} = \operatorname{median}\big(\{(\mathbf{j}_i^t)_z \mid t=1,\dots,T, i = 1,\dots,J\}\big),
\end{equation}
and if $|\tilde{z}| < \tau$ (we use $\tau = 0.5$ in practice), we translate all joints and associated Gaussians along the optical axis by a constant offset $\Delta z$:
\begin{equation}
    \mathbf{j}_i^{t\,'} = \mathbf{j}_i^t + \Delta z\,\mathbf{e}_z,\quad
    \Delta z =
    \begin{cases}
        1, & |\tilde{z}| < \tau,\\[2pt]
        0, & \text{otherwise,}
    \end{cases}
\end{equation}
where $\mathbf{e}_z = (0,0,1)^\top$. This deterministic depth-shift is applied consistently across all frames to preserve temporal continuity. Together with the fixed $(\mathbf{K}, \mathbf{W})$, it defines a canonical, front-view camera for all sequences.

\subsubsection{Neural Feature Modulator (NFM) Architecture}

\paragraph{Input Feature Vector.}
The NFM conditions appearance on local kinematics, enabling the renderer to highlight motion-salient regions adaptively. For each Gaussian $k$, we compute its aggregated position $\mathbf{p}_k^t$ and velocity $\mathbf{v}_k^t$ via the same LBS weights used for geometric deformation:
\begin{equation}
\mathbf{p}_k^t = \sum_{i=1}^J w_{k,i} \mathbf{j}_i^t, \qquad
\mathbf{v}_k^t = \sum_{i=1}^J w_{k,i} \dot{\mathbf{j}}_i^t.
\end{equation}
A lightweight appearance head $\text{MLP}_\text{app}: \mathbb{R}^d \to \mathbb{R}^4$ maps the learnable canonical feature $\mathbf{f}_k$ to a base RGBA value: $\text{RGBA}_k^\text{base} = \text{MLP}_\text{app}(\mathbf{f}_k)$. The NFM input is the concatenation $\mathbf{x}_k^t = [\mathbf{p}_k^t, \mathbf{v}_k^t, \text{RGBA}_k^\text{base}] \in \mathbb{R}^{10}$.

\paragraph{Network Architecture and Output.}
\revise{A} single-layer GRU (hidden size 10) models temporal dependencies across frames, producing $\mathbf{h}_k^t = \text{GRU}(\mathbf{x}_k^t, \mathbf{h}_k^{t-1})$. A two-layer MLP ($10 \to 64 \to 5$) with ReLU activation then predicts:
\begin{equation}
[\Delta \text{RGB}_k, \Delta \alpha_k, g_k] = \text{MLP}_\text{NFM}(\mathbf{h}_k^t),
\end{equation}
where $\Delta \text{RGB}_k \in \mathbb{R}^3$ and $\Delta \alpha_k \in \mathbb{R}$ are appearance residuals, and $g_k \in \mathbb{R}$ is a saliency logit. The modulated color and opacity are computed as:
\begin{align}
\mathbf{C}_k^\text{learned} &= \sigma(\text{RGB}_k^\text{base} + \Delta \text{RGB}_k), \\
\alpha_k^\text{learned} &= \sigma(\alpha_k^\text{base} + \Delta \alpha_k) \cdot \sigma(g_k),
\end{align}
where $\sigma(\cdot)$ denotes the sigmoid function. The saliency gate $\sigma(g_k)$ allows the network to suppress visually uninformative Gaussians (e.g., stationary joints) while amplifying those undergoing significant motion.

\paragraph{Depth-Color Blending.}
The final rendered color blends the NFM-modulated appearance with a depth-based pseudo-color visualization:
\begin{equation}
\mathbf{C}_k = (1 - \lambda)\, \mathbf{C}_k^\text{learned} + \lambda\, \mathbf{C}_k^\text{depth},
\end{equation}
where $\lambda = \sigma(\theta_\text{mix})$ is a learnable mixing weight and $\mathbf{C}_k^\text{depth}$ maps the Gaussian's depth to a triangular RGB colormap ($\text{red} \to \text{green} \to \text{blue}$ from near to far). This blending preserves spatial structure while allowing learned colors to encode semantic motion cues.

\subsection{Details of MQA, Disc-FT and CR-Distill}
\label{sec:suppl_prompt_design}

\begin{figure*}[!htbp]
	\centering 
	\begin{tabular}{c}		
		\includegraphics[width=\textwidth]{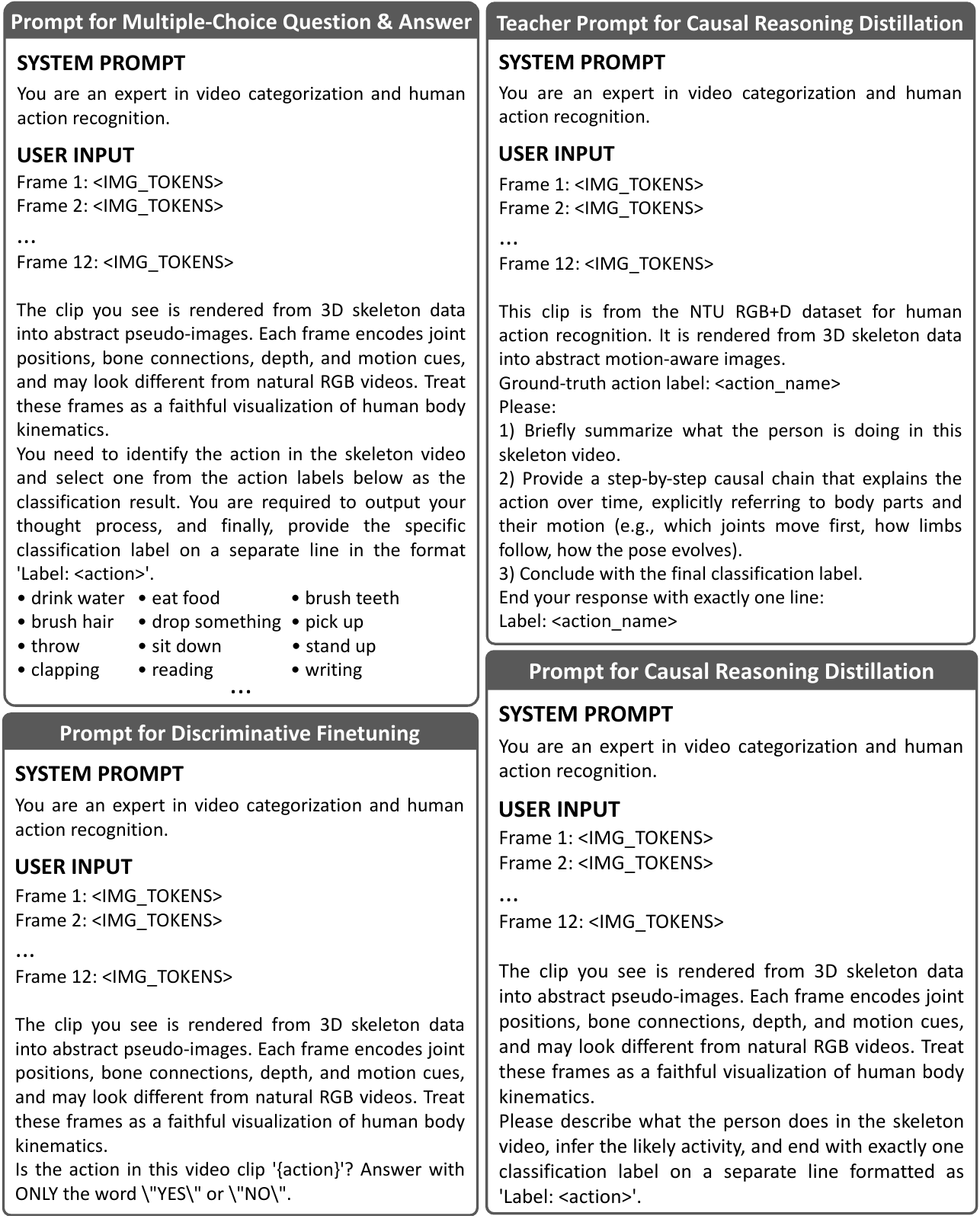}\\
	\end{tabular}%
	\caption{Prompt templates for MQA, Disc-FT, and CR-Distill. (Top-left) Prompt template for the MQA task. (Bottom-left) Prompt template for Disc-FT. (Top-right) Teacher prompt for CR-Distill. (Bottom-right) Student prompt for CR-Distill.}
    \label{fig:prompt_all}%
\end{figure*}

This section provides additional details on the three training tasks used in SkeletonLLM---the Multiple-Choice Question \& Answer (MQA) task, Discriminative Finetuning (Disc-FT), and Causal Reasoning Distillation (CR-Distill). We include the corresponding natural-language prompt templates, and the complete prompt texts are shown in \cref{fig:prompt_all}.

\subsubsection{Multiple-Choice Question \& Answer}
\label{sec:suppl_mqa_prompt}

We formulate open-vocabulary skeleton-based action recognition as a multiple-choice question answering task over a set of candidate action labels. For each clip, DrAction renders a sequence of $N_{\text{frames}}$ pseudo-images that are fed to the MLLM together with a text query listing the candidate classes and instructing the model to output a final decision in the format \texttt{Label: <action>}. During training, the candidate list contains only seen (training) classes; at test time, it is constructed from the corresponding evaluation split. The corresponding prompt template is illustrated in \cref{fig:prompt_all} (top-left).

We optimize the model by minimizing the negative log-likelihood of the ground-truth answer sequence, focusing on the tokens forming the final label line. At inference, we use greedy autoregressive decoding, selecting at each step the token with the highest predicted probability until termination. We then extract the action string following the \texttt{Label:} prefix via regex and apply standard text normalization (lowercasing, stripping leading/trailing whitespace) before comparing with the ground-truth label.

\subsubsection{Discriminative Finetuning}
\label{sec:suppl_discft_prompt}

Discriminative Finetuning (Disc-FT) is the second stage in our progressive training strategy. Its goal is to sharpen the decision boundaries between visually similar actions by teaching SkeletonLLM to answer targeted binary questions. Instead of selecting a label from a long candidate list, the model is asked whether a rendered clip depicts a specific action and must respond with a single ``YES'' or ``NO'' token. The exact prompt template is shown in \cref{fig:prompt_all} (bottom-left).

\begin{figure*}[!htbp]
	\centering 
	\begin{tabular}{c}		
		\includegraphics[width=\textwidth]{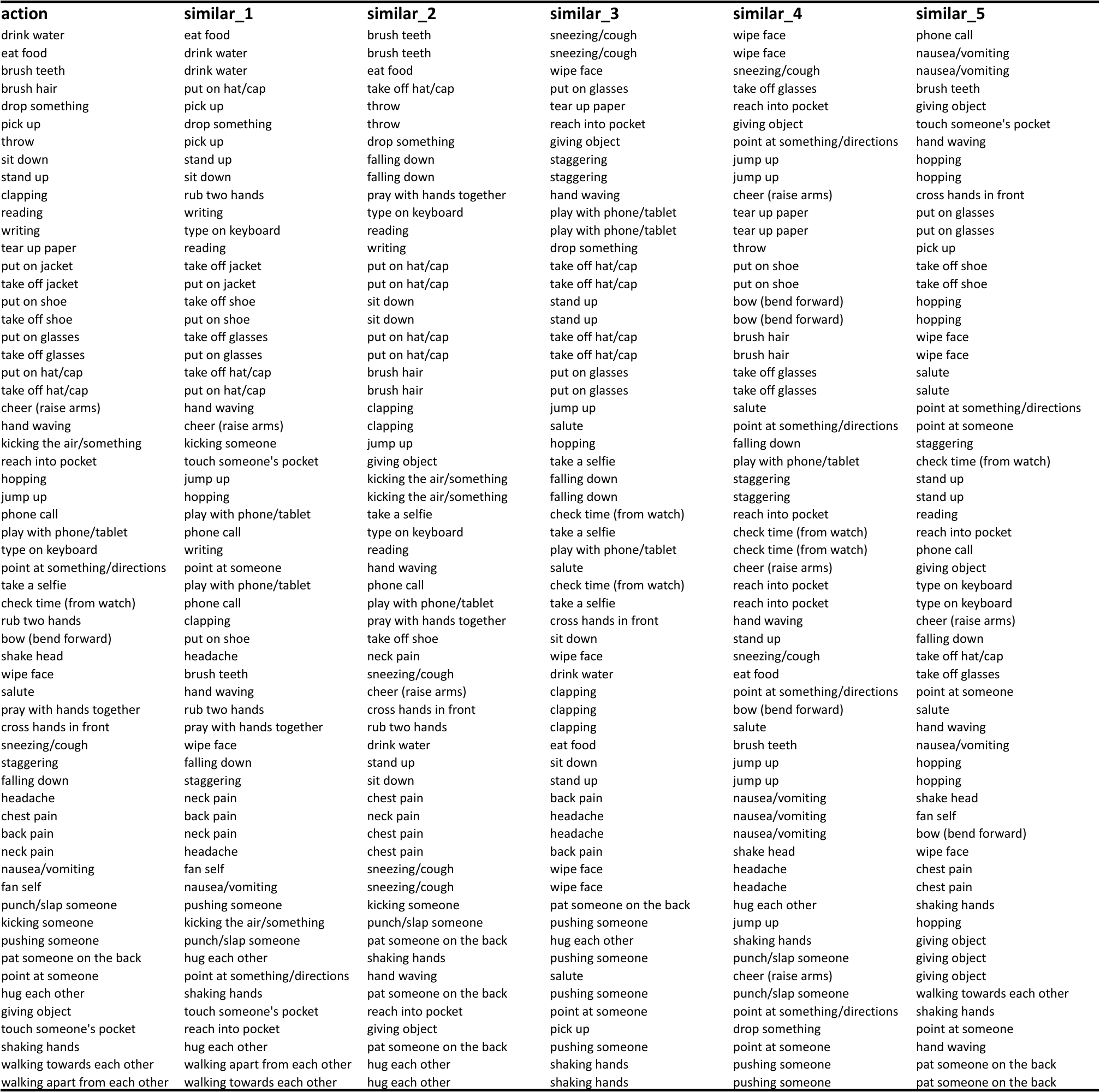}\\
	\end{tabular}%
	\caption{\revise{Top-5 MLLM-mined semantically similar actions for each class on NTU-60, used as hard-negative candidates in Disc-FT. Actions sharing similar body-part involvement form natural clusters (e.g., ``clapping,'' ``rub two hands,'' ``pray with hands together,'' ``high-five''); confusing pairs often differ only in subtle temporal or spatial cues (e.g., ``put on shoe'' vs.\ ``take off shoe,'' ``sit down'' vs.\ ``stand up'').}}\label{fig:ntu60_label}%
\end{figure*}

\begin{figure*}[!htbp]
	\centering 
	\begin{tabular}{c}		
		\includegraphics[width=\textwidth]{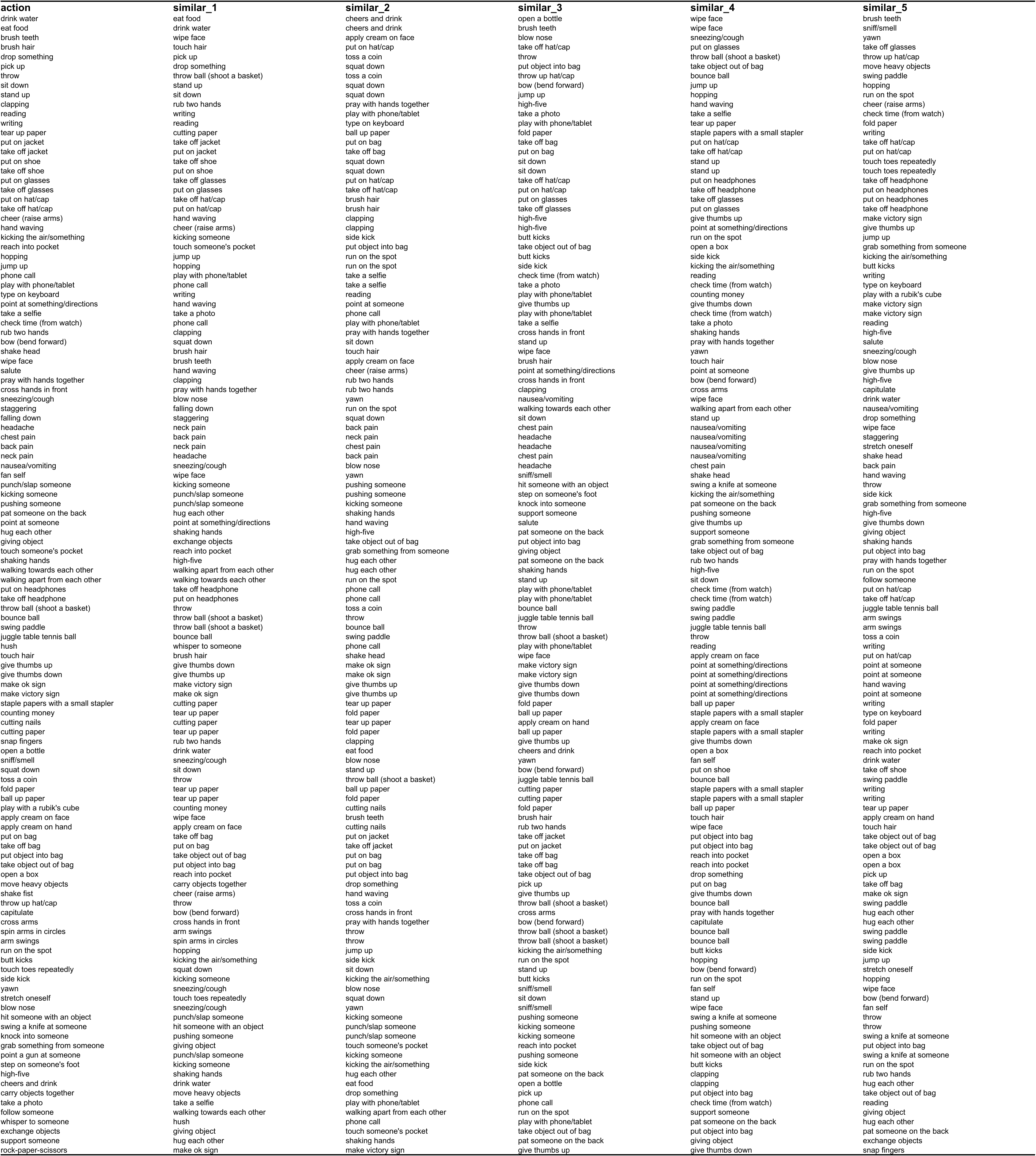}\\
	\end{tabular}%
	\caption{\revise{Top-5 MLLM-mined semantically similar actions for each class on NTU-120, used as hard-negative candidates in Disc-FT.}}\label{fig:ntu120_label}%
\end{figure*}

To focus learning on hard cases, Disc-FT operates on pairs of semantically similar actions. For each dataset and for every class name $y$, we query a strong MLLM \citep{Hurst2024GPT4oSC} with all other class names and ask it to rank which actions are most similar to $y$ in terms of body parts involved and motion patterns. We then retain the top-5 most similar actions as candidate negatives for $y$. The mined neighbors for NTU-60 and NTU-120 are visualized in \cref{fig:ntu60_label,fig:ntu120_label}. \revise{Several consistent patterns emerge: actions sharing similar body-part involvement form natural clusters (e.g., hand-centric actions such as ``clapping,'' ``rub two hands,'' ``hand waving,'' and ``cheer''), and many confusing pairs differ only in subtle temporal or spatial cues (e.g., ``put on shoe'' vs.\ ``take off shoe,'' ``bow'' vs.\ ``hopping''). These mined neighborhoods directly validate Disc-FT's design: the hardest recognition errors stem from semantically proximate categories that require fine-grained visual distinction.}

Disc-FT must also respect the open-vocabulary evaluation protocol. For a given split (e.g., the 48/12 split on NTU-60), only classes in the seen set are allowed to appear in training prompts. Thus, for each anchor class $y$ in the seen set, we intersect its MLLM-mined neighbor list with the seen classes of the current split and discard any neighbors that belong to the unseen set. Given a training clip $(\mathbf{V}, y)$ whose label $y$ is in the seen set, we construct Disc-FT samples based on this filtered neighbor set. For positive (``YES'') samples, we instantiate the template with the true label $y$, and the correct target token is ``YES''. For negative (``NO'') samples, we randomly select a label $z$ from the filtered candidate set of $y$ (with $z \neq y$) and ask the same question with the action replaced by $z$; the correct target token is ``NO''. We balance the training set such that the number of ``YES'' and ``NO'' examples is approximately 1:1.

\subsubsection{Causal Reasoning Distillation}
\label{sec:suppl_cr_prompt}

CR-Distill aims to inject explicit, temporally grounded reasoning into SkeletonLLM. Instead of supervising the model only with class labels, we use a stronger MLLM as a teacher to provide step-wise causal chains that describe how the action unfolds over time.

\begin{figure*}[!htbp]
	\centering 
	\begin{tabular}{c}		
			\includegraphics[width=\textwidth,height=0.86\textheight,keepaspectratio]{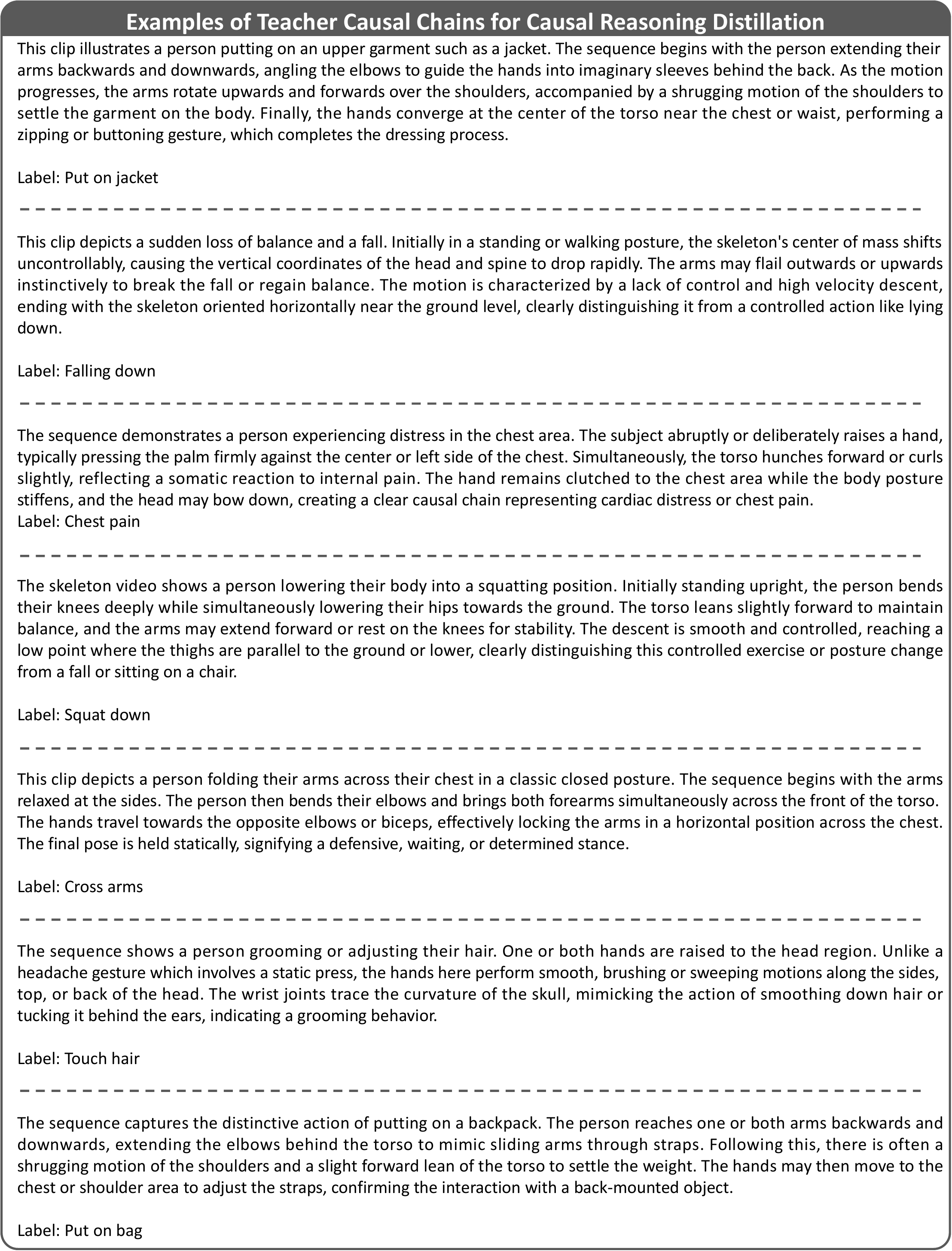}\\
	\end{tabular}%
	\caption{Examples of teacher-generated causal reasoning chains used in CR-Distill.}\label{fig:reason}%
\end{figure*}

We adopt GPT-4o \citep{Hurst2024GPT4oSC} as the teacher model. For each training clip, we first render the skeleton sequence into pseudo-image frames using the DrAction renderer after Stage~1 (Render Warm-up) and Stage~2 (Disc-FT) have been trained. These frames are fed to the teacher together with the prompt shown in \cref{fig:prompt_all} (top-right). The prompt states that the clip is rendered from 3D skeleton data, provides the ground-truth action label, and asks the teacher to (i) briefly summarize the action, (ii) produce a step-by-step causal chain that explicitly refers to body parts and their motion, and (iii) end with a final line \texttt{Label: <action\_name>}. Representative examples of such teacher-generated causal chains are illustrated in \cref{fig:reason}.

Given the same rendered frames, SkeletonLLM is queried with the student prompt in \cref{fig:prompt_all} (bottom-right), which explains that the input frames are abstract pseudo-images encoding joint positions, bone connections, depth, and motion cues, and asks the model to describe the action and output a final label line formatted as \texttt{Label: <action>}. The student is not given the ground-truth label in the prompt; instead, it is trained to reproduce the full teacher response, including both the causal rationale and the terminal \texttt{Label:} line. We minimize an auto-regressive cross-entropy loss over the teacher token sequence, masking out the prompt tokens. During this stage, we update the parameters of DrAction, the vision-to-language projector, and the LoRA adapters of the language model, encouraging the whole pipeline to encode motion cues in a way that supports detailed, body-part-specific reasoning.

\section{Design Rationale: Why 3DGS + LBS}
\label{sec:design_rationale}

We employ 3D Gaussian Splatting driven by Linear Blend Skinning to address the specific constraints of skeletal data: the need for structural preservation, kinematic validity, format agnosticism, and differentiable optimization.

3DGS offers a discrete set of primitives that naturally map to the nodes and edges of a skeleton graph. Unlike implicit neural fields (e.g., NeRF) which represent geometry continuously, 3DGS allows us to explicitly attach visual elements to specific body parts. This discrete correspondence is vital for part-aware reasoning, as it ensures that the learned visual features are directly grounded in the underlying skeletal topology. Crucially, 3DGS primitives can be dynamically instantiated based on the input skeleton's joint count and connectivity, enabling format-agnostic rendering across heterogeneous skeleton topologies (e.g., Kinect, MoCap, 2D poses).

LBS acts as a robust kinematic regularizer. By binding the Gaussian primitives to joint transforms, LBS ensures that the rendered motion remains physically plausible and temporally consistent. Without this constraint, optimizing primitives freely could lead to artifacts where visual elements detach from the skeletal structure, breaking the semantic link between motion and appearance. LBS also provides a principled mechanism for handling arbitrary skeleton formats: the blending weights are computed based on the input skeleton's adjacency structure rather than a fixed template.

This combination enables a design that is both structured and expressive. While LBS enforces geometric consistency, the learnable parameters of 3DGS (e.g., opacity, color) provide the degrees of freedom needed to encode semantic information. Through end-to-end gradients, the model can learn to highlight salient motion cues (such as a moving limb) or encode velocity into color, creating a visual language optimized for understanding rather than mere reconstruction. The format-agnostic nature of this design is central to achieving universal skeleton understanding across diverse data sources.

\section{Future Work}
\label{sec:future_work}

SkeletonLLM takes a first step toward enabling MLLMs to operate on sparsely structured, non-visual signals by translating skeleton sequences into a task-optimized visual language. Natural extensions include applying DrAction-style differentiable rendering to other structured modalities (e.g., LiDAR point clouds, object trajectories), and exploiting the differentiability for motion generation---optimizing skeleton sequences to match textual descriptions for text-to-motion synthesis.

Another avenue concerns long-horizon, compositional action understanding. Our current evaluation focuses on short clips depicting atomic actions, yet real-world scenarios often involve minute-scale sequences with multiple sub-actions and temporal dependencies, requiring hierarchical temporal modeling and efficient memory mechanisms.

Finally, improving visual token efficiency remains an open challenge. As shown in \cref{tab:resolution}, reducing resolution from $448 \times 448$ to $224 \times 224$ leads to $\sim$5\% performance degradation. More fundamentally, rendered skeleton images are inherently sparse: the human figure occupies only a small fraction of pixels, while the majority of the frame consists of uninformative black background. This sparsity suggests that current dense visual encoding is suboptimal. Potential solutions include sparse attention mechanisms that focus computation on skeleton-occupied regions, or hybrid representations that combine compact rendered patches with numerical joint encodings to reduce redundant pixel processing.

\end{document}